\documentclass[10pt,twocolumn,letterpaper]{article}

\usepackage{cvpr}              
\definecolor{cvprblue}{rgb}{0.21,0.49,0.74}
\usepackage[pagebackref,breaklinks,colorlinks,allcolors=cvprblue]{hyperref}

\usepackage{algorithm}      
\usepackage{amsmath}        
\usepackage{amssymb}        
\usepackage{bm}             
\usepackage{multirow}
\usepackage{longtable}
\usepackage{booktabs} 
\usepackage{paracol}
\usepackage{hyperref}   

\usepackage{algorithmic}  

\def\paperID{4428} 
\def\confName{CVPR}
\def\confYear{2026}

\title{No Way To Steal My Face: Proactive Defense Against Identity-Preserving Personalized Generation}

\author{
\parbox{\linewidth}{\centering
Lizhi Xiong\textsuperscript{1}, 
Jun Li\textsuperscript{1}, 
Ziqiang Li\textsuperscript{1}\thanks{Corresponding author.}\ ,\ 
Weiwei Jiang\textsuperscript{1}, 
Zhangjie Fu\textsuperscript{1} 
\\[6pt] 
\small
\textsuperscript{1}Nanjing University of Information Science and Technology
\\[6pt]
\small
\ttfamily
lzxiong16@163.com, lijuun@yeah.net, iceli@mail.ustc.edu.cn
\\
weiwei.jiang@nuist.edu.cn, fzj@nuist.edu.cn
}
}

\begin{document}
\maketitle

\begin{abstract}
Recent advances in diffusion models have enabled high-fidelity, identity-preserving image generation for personalized applications such as digital avatars and virtual try-on systems. However, their reliance on sensitive facial reference images raises growing privacy concerns. Existing defense mechanisms are primarily designed for training-based personalization and struggle to generalize to emerging training-free approaches, due to fundamental differences in their identity integration paradigms. To bridge this gap, we propose \textbf{IDGuardian}—the first generalizable and model-agnostic identity protection framework capable of defending against both training-based and training-free personalization methods. IDGuardian abstracts the personalization process into two critical stages: identity extraction and identity injection. It then introduces crafted adversarial perturbations to simultaneously disrupt both stages. Specifically, it degrades the identity features extracted by external encoders and establishes an adversarial conceptual bridge that misdirects the generative trajectory away from the target identity. Extensive experiments show that IDGuardian effectively protects identity across various personalization pipelines and model architectures, while remaining robust to post-processing, adaptive attacks, and cross-dataset generalization. Our code is available at \url{https://github.com/OpenAscent-L/IDGuardian.git}.
\end{abstract}

\section{Introduction}
\label{sec:intro}

\begin{figure}[htbp]
    \centering
    \includegraphics[width=0.48\textwidth]{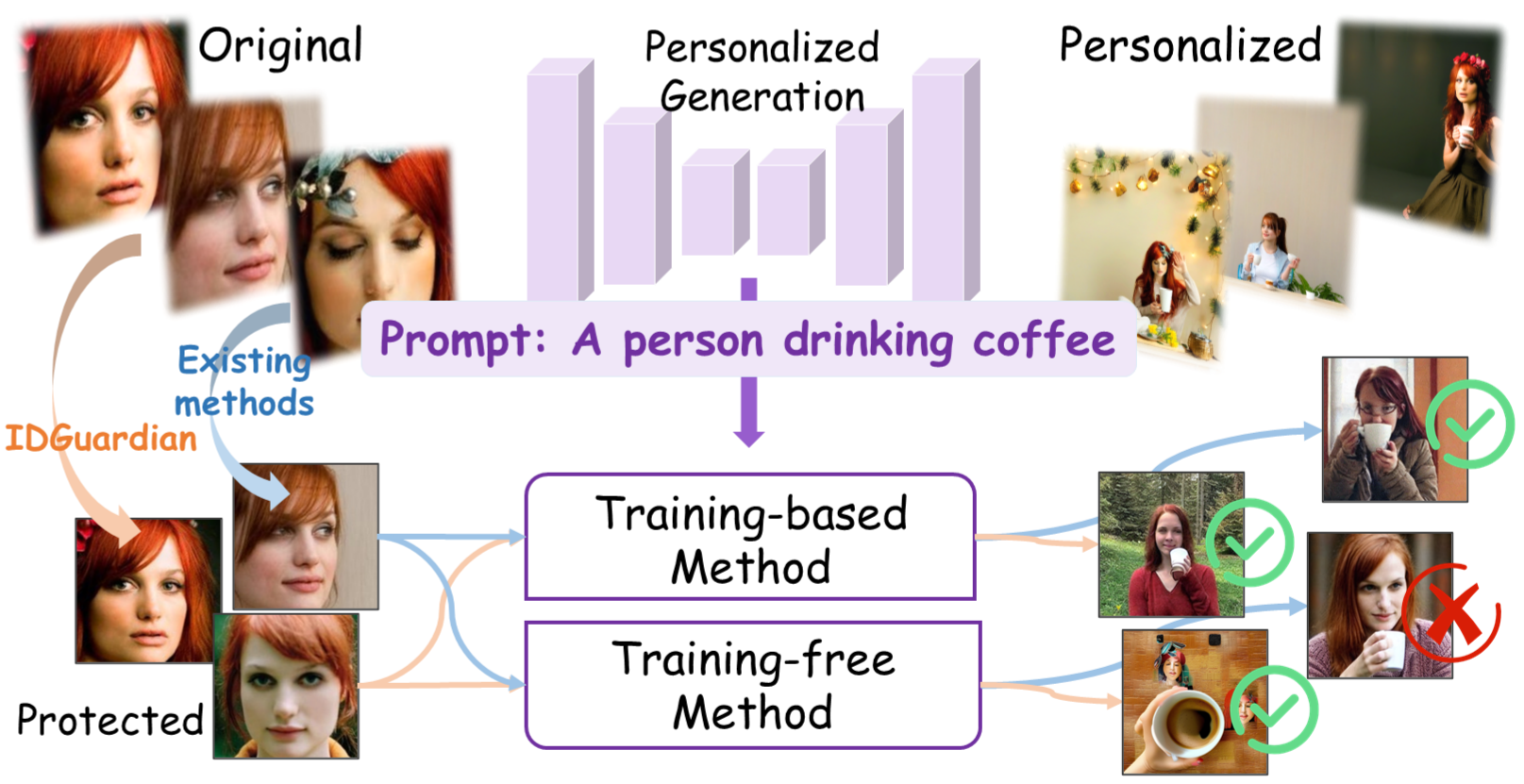}
    \caption{Existing defense methods become ineffective in training-free personalization due to paradigm discrepancies and fusion diversity.}
    \label{fig:paradigm_diff}
\end{figure}

Recent advances in diffusion models have significantly improved high-fidelity, identity-preserving image generation \cite{rombach2022high, li2025comprehensive}. By conditioning on facial reference images, these models can produce photorealistic portraits that accurately retain individual identities \cite{li2023blip, rout2024rb, valevski2023face0}. Existing identity-preserving personalization methods generally fall into two categories: \textbf{training-based} and \textbf{training-free} approaches.
Training-based methods \cite{ruiz2023dreambooth, kumari2023multi, hu2022lora} fine-tune pre-trained diffusion models on a small set of user-specific images, effectively capturing detailed identity features. However, they require significant computational resources, careful hyperparameter tuning, and GPU support, which limits their scalability in practice.
In contrast, training-free methods \cite{ye2023ip, wu2024infinite, li2024photomaker} leverage identity encoders pre-trained on large-scale datasets to extract semantic facial representations in a zero-shot manner during inference.  These methods enable fast, scalable, and user-friendly personalization, making them popular in real-world applications due to their efficiency and low deployment cost.

Despite the impressive progress in personalized image generation enabled by diffusion models, growing privacy concerns have arisen due to their reliance on sensitive facial reference images, which can be misused for identity reconstruction or impersonation. To address these risks, recent proactive defenses have explored perturbing input images to disrupt the generation of personalized content \cite{wang2024edit}. These methods \cite{van2023anti, shan2023glaze, huang2021unlearnable, sunpretender} typically adopt training-based personalization as a proxy task, optimizing adversarial perturbations to hinder the model’s ability to encode identity-specific features.
However, their effectiveness degrades markedly when applied to training-free personalization, as illustrated in Figure~\ref{fig:paradigm_diff}, due to two core challenges arising from differences in personalization paradigms:
\textbf{i) Personalized Pipeline Discrepancy.}  Training-based approaches internalize identity features through model fine-tuning, whereas training-free methods inject identity embeddings via pre-trained encoders without modifying model parameters. This structural distinction renders perturbations optimized under the training-based setting ineffective in training-free scenarios.
\textbf{ii) Information Fusion Diversity.} Training-free pipelines exhibit considerable heterogeneity in encoder architectures and identity-semantics fusion strategies, lacking a unified framework. This diversity poses a significant obstacle to developing generalizable defenses, leaving critical privacy risks unresolved.

To bridge this gap, we propose \textbf{IDGuardian}, a generalizable and model-agnostic defense mechanism designed to protect identity privacy across both training-based and training-free personalization paradigms.\footnote{IDProtector~\cite{Song_2025_CVPR} addresses training-free scenarios by maximizing identity inconsistency across multiple encoders, thereby disrupting feature alignment without relying on diffusion gradients. However, its reliance on specific architectural assumptions limits its applicability to training-based methods and encoder-agnostic personalization.}
Despite the structural differences among existing personalization pipelines, \textbf{IDGuardian} mitigates identity leakage by abstracting the generation process into two fundamental stages: identity extraction and identity injection. It then formulates a dual-perspective adversarial strategy that disrupts identity learning through the following mechanisms:
\textbf{i) Cross-Encoder Identity Field Confounding.}
IDGuardian enforces intentional dissimilarity between identity features extracted from the protected and original images, hindering malicious exploitation of the reference identity in downstream generation. 
\textbf{ii) Guidance-Flow Identity Deflection.}
By interpreting identity-preserving personalization as a conceptual bridge from an unconditional to an identity-conditioned distribution, IDGuardian establishes an adversarial conceptual bridge that redirects the diffusion sampling trajectory away from the clean identity manifold, thereby impairing identity consistency during synthesis.

To evaluate the effectiveness and robustness of \textbf{IDGuardian}, we conduct extensive experiments across diverse personalization pipelines, including training-based and training-free methods, as well as diffusion- and flux-based generators, and observe strong transfer from a single white-box training setup to unseen black-box systems. The protection remains effective under post-processing, adaptive attacks, and cross-dataset transfer. Ablation studies verify the contribution of each component, while both human and perceptual evaluations confirm reliable identity protection without compromising usability. 
Our contributions can be summarized into three-fold:
\begin{itemize}
\item We introduce the first proactive identity protection framework that defends against both training-free and training-based personalization methods, effectively mitigating malicious identity misuse. Our approach fills a critical gap in existing defenses by demonstrating dual adaptability across personalization paradigms.
\item We propose \textbf{IDGuardian}, which abstracts identity-preserving generation into identity extraction and injection. Through Cross-Encoder Identity Field Confounding and Guidance-Flow Identity Deflection, it disrupts feature alignment and redirects diffusion trajectories, mitigating identity preservation across personalization pipelines.
\item Extensive experiments show that IDGuardian effectively protects identity across diverse personalization methods, including generators based on diffusion and flux. Trained on a single white-box diffusion model, it generalizes to unseen black-box pipelines and datasets, and remains robust against post-processing and adaptive attacks.
\end{itemize}

\section{Related Work}
\label{sec:Related work}

\subsection{Identity-Preserving
Personalized Generation}

Identity-preserving personalized generation aims to synthesize images that faithfully retain the unique identity of a given subject \cite{chen2019pog, shi2024instantbooth}. Recent advances in diffusion-based models have significantly enhanced the fidelity and identity consistency of generated images \cite{podell2023sdxl, he2024uniportrait}. Existing methods can be broadly categorized into two classes: \textbf{training-based} and \textbf{training-free}. Training-based methods—such as DreamBooth \cite{ruiz2023dreambooth}, CustomDiffusion \cite{kumari2023multi}, Textual Inversion \cite{gal2022image}, and LoRA-based fine-tuning \cite{hu2022lora}—adapt pre-trained diffusion models to new identities by fine-tuning them on a small set of reference images. These approaches often yield high-quality personalization and effectively preserve detailed identity cues, but they are computationally intensive and sensitive to hyperparameter settings, limiting their scalability and usability in practical scenarios.

To address these challenges, training-free approaches have emerged as lightweight and efficient alternatives \cite{cui2024idadapter, guo2024pulid}. These methods typically use external identity encoders to extract semantic facial representations zero-shot, injecting them into the diffusion process via various fusion strategies without modifying model parameters \cite{zhou2024storymaker}. The key difference among recent training-free methods lies in how identity is integrated into generation \cite{li2025comprehensive}. For example, PhotoMaker \cite{li2024photomaker} uses a Direct Stack strategy to encode multiple references into a unified identity embedding replacing class-specific tokens. IP-Adapter \cite{ye2023ip} and InstantID \cite{wang2024instantid} employ Cross-Attention to inject identity features alongside text-based attention. Infinite-ID \cite{wu2024infinite} extends this with a Mixed Attention scheme combining cross- and self-attention. In contrast, KV-Edit \cite{zhu2025kv} relies solely on Self-Attention, caching background key-value pairs and merging them with foreground queries for identity-aware editing.

\subsection{Proactive Defense for Personalized Generation}

With the rapid proliferation of personalized generative models, concerns about privacy leakage and the unauthorized use of personal images have spurred growing interest in developing effective defense mechanisms \cite{ye2023duaw, song2024anti, liu2023diffprotect, li2025artificial}. Among these, active defense strategies aim to proactively perturb user-provided reference images to prevent downstream models from accurately capturing and reproducing identity features. Recent advances have introduced a variety of active defense techniques designed to disrupt the fine-tuning processes of generative models \cite{madry2017towards, an2024sd4privacy, he2024diff}.
For instance, Anti-DreamBooth \cite{van2023anti} applies PGD-based perturbations \cite{madry2017towards} to interfere with the fine-tuning pipeline of models such as DreamBooth. AdvDM \cite{liang2023adversarial} leverages Monte Carlo estimation to optimize perturbations in the latent space of diffusion models, preventing the extraction of features from protected artworks. SimAC \cite{wang2024simac} enhances perturbation effectiveness through greedy timestep selection and a feature-level interference loss tailored to diffusion dynamics. In contrast, PersGuard \cite{liu2025persguard} implants backdoor triggers into pre-trained models to selectively disable personalization for protected inputs. Additionally, ACE \cite{zheng2023understanding} proposes an unified adversarial framework that jointly optimizes encoder error, reverse process error, and fine-tuning error, employing target-guided optimization to balance conflicts among these objectives.

However, the effectiveness of existing defenses significantly diminishes in training-free settings, primarily due to Personalized Pipeline Discrepancy and Information Fusion Diversity inherent in personalization methods. As training-free pipelines gain traction for their scalability and ease of deployment, there is an urgent need for universal defense strategies that can robustly counteract both training-based and training-free personalization approaches.
In contrast to training-based pipelines, effective defense against training-free personalization remains insufficiently explored, particularly in ways that can generalize across both paradigms.
Concurrent with our work, IDProtector \cite{Song_2025_CVPR} addresses identity leakage by aligning features across identity encoders used in various personalization pipelines. Although effective in training-free scenarios, it relies on multiple encoder-specific features tied to specific models, limiting its generalization to training-based pipelines and introducing considerable architectural complexity. These limitations underscore the need for more efficient, generalizable, and model-agnostic defense mechanisms.

\begin{figure*}[htbp]
    \centering
    \includegraphics[width=1\textwidth]{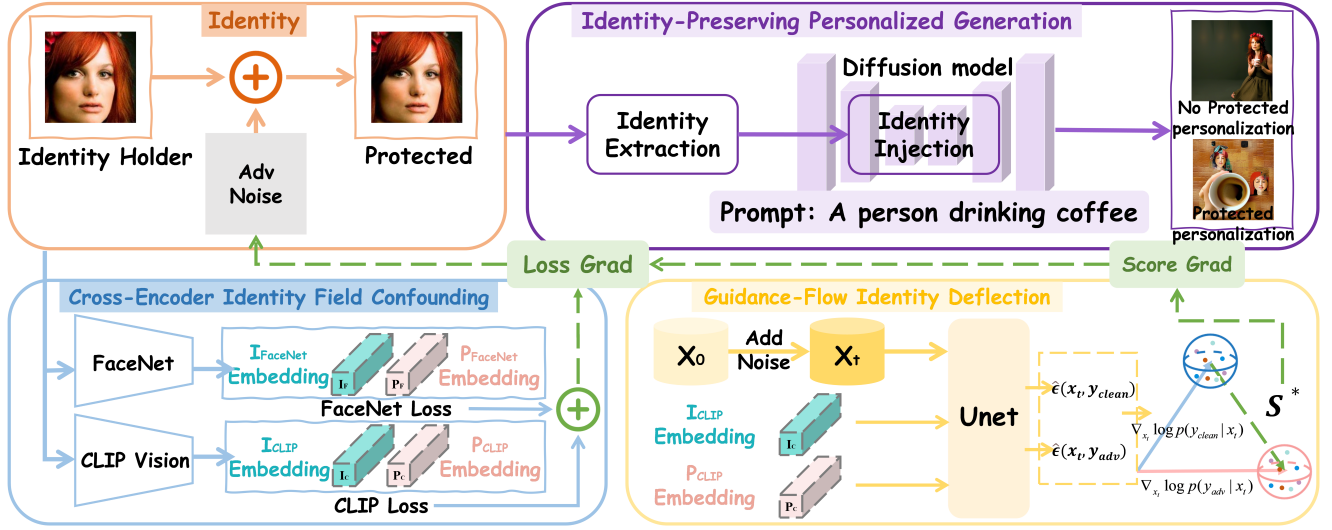}
    \caption{\textbf{Overview of IDGuardian.} The framework centers on two complementary stages of personalization: \textbf{identity extraction} and \textbf{identity injection}. To disrupt identity extraction, \textbf{Cross-Encoder Identity Field Confounding} decorrelates visual identity embeddings across feature spaces. To weaken identity injection, \textbf{Guidance-Flow Identity Deflection} redirects the denoising trajectory toward identity-neutral regions. Identity loss is computed using embeddings from the original (\( \mathbf{I}_{\text{FaceNet/CLIP}} \)) and protected (\( \mathbf{P}_{\text{FaceNet/CLIP}} \)) images, while distributional deviation is induced via adversarial conceptual bridge during the denoising process.}
    \label{fig:IDGuardian}
\end{figure*}

\section{Methods}
\label{sec:Methods}

\subsection{Threat Model}

We consider a real-world identity abuse scenario involving three roles: the \textbf{identity holder} (victim), the \textbf{generative user} (attacker), and the \textbf{protection mechanism} (defender) deployed by the identity holder.

\noindent\textbf{Identity Holder (Victim).} 
The identity holder shares facial images on public platforms. These images may be harvested and exploited to synthesize novel identity-consistent content without consent. To mitigate such privacy risks, the identity holder proactively applies subtle, imperceptible perturbations that preserve visual realism while reducing extractability of identity-relevant features.

\noindent\textbf{Generative User (Attacker).}
The attacker uses personalized generation techniques, often diffusion-based, to synthesize images that preserve the victim’s identity, enabling scalable and high-fidelity reproduction.

\noindent\textbf{Protection Mechanism (Defender).}  
To prevent unauthorized identity preservation, the defender perturbs the input image. Operating in a restricted access setting, the defender has no knowledge of the specific personalized generation techniques employed by the attacker and is limited to modifying the input image alone. 
Given any personalized identity-preserving generator \(G(\cdot)\) and input image \(x\), the defender seeks to optimize a perturbation \(\delta\) such that \(x + \delta\) reduces identity consistency in the generated output. The objective is to ensure that \(G(x + \delta)\) deviates in identity from \(x\), thereby diminishing the effectiveness of personalized generation in preserving the victim’s identity. The optimization problem is formulated as:
\[
\min_{\delta} \text{SIM}\bigl(\text{ID}(x), \text{ID}(G(x + \delta))\bigr) \quad \text{s.t.} \quad \|\delta\|_\infty < \epsilon,
\]
where $\text{ID}(\cdot)$ denotes the identity feature extractor, $\text{SIM}(\cdot, \cdot)$ quantifies the similarity between identities, and \(\epsilon\) constrains the magnitude of the perturbation to ensure imperceptibility.

\subsection{Overall Method}

Identity-preserving personalized generation methods typically consist of two critical stages: identity extraction and identity injection. In the first stage, an identity extractor derives identity features from a reference image. In the second, these features are fused with semantic content via identity injection techniques to produce personalized outputs. As depicted in the upper part of Figure~\ref{fig:IDGuardian}, this pipeline can be abstracted into a two-stage process. Leveraging this structure, we propose \textbf{IDGuardian}, a defense framework that protects a reference identity by independently disrupting both the identity extraction and injection stages.
\begin{itemize}
  \item \textbf{Cross-Encoder Identity Field Confoundin:} IDGuardian enforces the identity features extracted from the protected image to be intentionally dissimilar from those of the original image, thereby hindering the malicious use of the reference identity in subsequent generative processes.
  \item \textbf{Guidance-Flow Identity Deflection:} IDGuardian establishes an adversarial conceptual bridge by interpreting identity-preserving personalization as a conceptual bridge from an unconditional to a target identity-conditioned distribution. This redirects the diffusion sampling trajectory away from the clean identity manifold, impairing identity consistency during synthesis.

\end{itemize}

By jointly optimizing perturbations at both stages, IDGuardian produces a final adversarial perturbation $\delta$ that remains imperceptible to human observers while substantially degrading identity consistency in the generated outputs.

\subsection{Cross-Encoder Identity Field Confounding}

To disrupt identity extraction, we explicitly minimize the identity similarity between the clean image \(x\) and its adversarially perturbed counterpart \(x + \delta\) within the embedding spaces of two complementary identity encoders. Employing dual encoders enables the capture of richer and more comprehensive identity representations, mitigating the limitations of using a single encoder, which may struggle to generalize across diverse image domains and identity variations. Specifically, we adopt FaceNet~\cite{schroff2015facenet} and CLIP~\cite{radford2021learning} as complementary identity extractors to enhance robustness of the identity loss. FaceNet is tailored for extracting facial identity features, while CLIP’s image encoder captures high-level semantic information from visual data, providing a broader and more nuanced representation of identity.

The identity loss is defined as a weighted sum of cosine similarities computed in the respective embedding spaces:
\begin{equation}
\mathcal{L}_{\text{ID}} = \mathcal{L}_{\text{FaceNet}} + \mathcal{L}_{\text{CLIP}},
\end{equation}
where each term measures the cosine similarity between features extracted from the clean and perturbed images:
\begin{equation}
\mathcal{L}_{\text{FaceNet}} = \cos\big(f_{\text{FaceNet}}(x), f_{\text{FaceNet}}(x + \delta)\big),
\end{equation}
\begin{equation}
\mathcal{L}_{\text{CLIP}} = \cos\big(f_{\text{CLIP}}(x), f_{\text{CLIP}}(x + \delta)\big),
\end{equation}
where \(f_{\text{FaceNet}}(\cdot)\) and \(f_{\text{CLIP}}(\cdot)\) denote the feature extraction functions that map input images to their respective identity embedding spaces.

\subsection{Guidance-Flow Identity Deflection}

To weaken identity injection during diffusion generation, we interpret identity-preserving personalization as a conceptual bridge represented by a structured shift in the score function toward the target identity manifold. Building on this, we establish an adversarial conceptual bridge that diverts the generative trajectory from the original manifold, disrupting identity preservation. The subsequent sections detail score-based foundations, the conceptual bridge perspective, and our adversarial conceptual bridge design.

\noindent\textbf{Score-Based Foundations in Diffusion.}
Diffusion models synthesize data by iteratively denoising a Gaussian noise sample through a learned reverse Markov process. During each timestep $t$, the model learns to approximate the gradient of the data log-probability—referred to as the \textit{score function}, denoted by $\nabla_{\mathbf{x}_t} \log p_\theta(\mathbf{x}_t)$—which steers the noisy input \( \mathbf{x}_t \) toward regions of higher data likelihood. As the exact score function is intractable to compute, diffusion models instead learn to predict the noise component \(\boldsymbol{\epsilon}_\theta(\mathbf{x}_t, t)\) introduced during the forward process. This leads to the following widely used approximation~\cite{ho2022classifier, song2020score, sohl2015deep, mcallister2024rethinking, liu2023more}:
\begin{equation}
\nabla_{\mathbf{x}_t} \log p_\theta(\mathbf{x}_t) \approx - \frac{1}{\sqrt{1 - \bar{\alpha}_t}} \cdot \boldsymbol{\epsilon}_\theta(\mathbf{x}_t, t),
\end{equation}
where \(\bar{\alpha}_t = \prod_{s=1}^t \alpha_s\) denotes the cumulative noise schedule up to timestep $t$.
In conditional generation settings, such as identity- or text-conditioned synthesis, the score function incorporates the conditioning variable \( \mathbf{y} \), yielding the approximation:
\begin{equation}
\nabla_{\mathbf{x}_t} \log p_\theta(\mathbf{x}_t \mid \mathbf{y}) \approx - \frac{1}{\sqrt{1 - \bar{\alpha}_t}} \cdot \boldsymbol{\epsilon}_\theta(\mathbf{x}_t, t , \mathbf{y}),
\end{equation}
where \( \boldsymbol{\epsilon}_\theta(\mathbf{x}_t, t, \mathbf{y}) \) denotes the predicted noise under the conditioning of \( \mathbf{y} \).

\noindent\textbf{Analyzing Identity-Preserving Personalization through the Lens of the Score Function as a Conceptual Bridge.}
Inspired by distributional transport frameworks such as Dual Diffusion Implicit Bridges~\cite{su2022dual}, 
we view identity-preserving personalization through the lens of a conceptual bridge. 
Grounded in the score-based foundation of diffusion models, this bridge can be interpreted as linking an unconditional to a conditional generative process, guiding generation toward the target identity $y$. 
This behavior can be modeled by maximizing the conditional likelihood \( \log p(y \mid \mathbf{x}_t) \), with its gradient reflecting how the identity $y$ influences the generative trajectory.
Applying Bayes' rule, the gradient can be decomposed as:
\begin{equation}
\nabla_{\mathbf{x}_t} \log p(y \mid \mathbf{x}_t) = \nabla_{\mathbf{x}_t} \log p(\mathbf{x}_t \mid y) - \nabla_{\mathbf{x}_t} \log p(\mathbf{x}_t),
\end{equation}
where the first term corresponds to the conditional likelihood, and the second term reflects the unconditional prior.

\begin{figure}[htbp]
    \centering
    \includegraphics[width=0.47\textwidth]{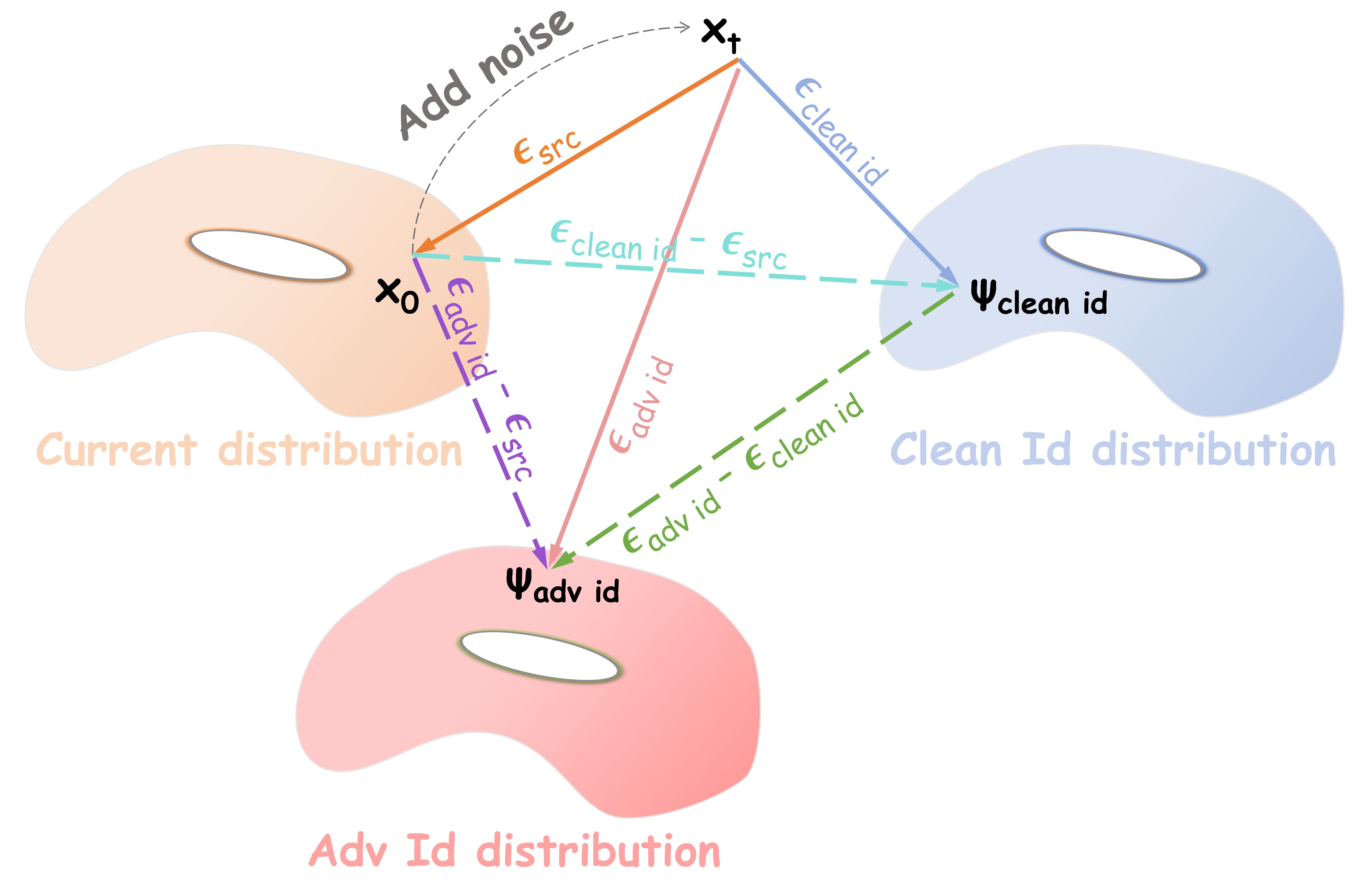}
    \caption{Visualization of \textbf{Conceptual Bridges} Constructed via Noise Prediction Differences in Diffusion Sampling.}
    \label{fig:DistributionShift}
\end{figure}

Following prior study~\cite{song2019generative}, this gradient difference \( \nabla_{\mathbf{x}_t} \log p(y \mid \mathbf{x}_t\)) can be approximated using the predicted noise of the model as:
\begin{equation}
\begin{split}
\nabla_{\mathbf{x}_t} \log p(y \mid \mathbf{x}_t) &\approx 
- \frac{1}{\sqrt{1 - \bar{\alpha}_t}} \left( {\epsilon}_\theta(\mathbf{x}_t, t, y) - {\epsilon}_\theta(\mathbf{x}_t, t, \emptyset) \right),
\end{split}
\end{equation}
where \( \epsilon_\theta(\mathbf{x}_t, t, y) \) represents predicted noise conditioned on the target identity, corresponding to the blue arrow \( \epsilon_{\text{clean id}} \) in Figure~\ref{fig:DistributionShift}. In contrast, \( \epsilon_\theta(\mathbf{x}_t, t, \emptyset) \) denotes the unconditional noise prediction, illustrated by the orange arrow \( \epsilon_{\text{src}} \). 
The difference between these two predictions (indicated by the light cyan dashed arrow) captures how identity conditioning alters the denoising direction, effectively steering the generative trajectory toward the target identity manifold~\cite{mcallister2024rethinking, hertz2023delta}.

This identity-induced distributional shift can be interpreted as a structured perturbation of the score function. Inspired by the analogy to distributional transport, we refer to this shift as a conceptual bridge, highlighting its role in connecting the source and identity-conditioned distributions in the generative process.

\noindent\textbf{Score-Based Adversarial Conceptual Bridge.}
Building on the conceptual bridge perspective of identity-preserving personalization, we introduce a parallel adversarial conceptual bridge that reorients the generative trajectory away from the clean identity manifold and toward a strategically designed adversarial identity distribution. Unlike its identity-preserving counterpart, this bridge deliberately disrupts identity consistency rather than reinforcing alignment. 

Formally, this adversarial conceptual bridge is constructed through the difference between the conditional gradients corresponding to adversarial and clean identity:
\begin{equation}
S^* = \nabla_{\mathbf{x}_t} \log p(y_{\text{adv}} \mid \mathbf{x}_t)
     - \nabla_{\mathbf{x}_t} \log p(y_{\text{clean}} \mid \mathbf{x}_t),
\end{equation}
where \( \nabla_{\mathbf{x}_t} \log p(y_{\text{adv}} \mid \mathbf{x}_t) \) can be approximated by the difference \( \epsilon_\theta(\mathbf{x}_t, t, y_{\text{adv}}) - \epsilon_\theta(\mathbf{x}_t, t, \emptyset) \), shown as the purple dashed arrow in Figure~\ref{fig:DistributionShift}, with \( y_{\text{adv}} \) extracted from the protected image, steering generation toward the adversarial identity manifold.
Concretely,
\begin{equation}
\begin{split}
S^* &\approx -\frac{1}{\sqrt{1 - \bar{\alpha}_t}} \cdot \left( \epsilon_\theta(\mathbf{x}_t, t, y_{\text{adv}}) - \epsilon_\theta(\mathbf{x}_t, t, \emptyset) \right) \\
&\quad + \frac{1}{\sqrt{1 - \bar{\alpha}_t}} \cdot \left( \epsilon_\theta(\mathbf{x}_t, t, y_{\text{clean}}) - \epsilon_\theta(\mathbf{x}_t, t, \emptyset) \right) \\
&= -\frac{1}{\sqrt{1 - \bar{\alpha}_t}} \cdot \left( \epsilon_\theta(\mathbf{x}_t, t, y_{\text{adv}}) - \epsilon_\theta(\mathbf{x}_t, t, y_{\text{clean}}) \right),
\end{split}
\end{equation}
where \( \epsilon_\theta(\mathbf{x}_t, t, y_{\text{adv}}) \) and \( \epsilon_\theta(\mathbf{x}_t, t, y_{\text{clean}}) \) denote predicted noise under adversarial and clean identity conditions, corresponding to the red and blue arrows \( \epsilon_{\text{adv id}} \) and \( \epsilon_{\text{clean id}} \) in Figure~\ref{fig:DistributionShift}. 
This adversarial conceptual bridge, shown as the green dashed arrow, suppresses identity preservation by steering the sample path to lower-likelihood regions under \(y_{\text{clean}}\). It integrates into the model’s scoring mechanism, ensuring temporal consistency and underpinning our defense.

\subsection{Perturbation Optimization}

We update the adversarial perturbation by integrating gradients derived from both the identity loss \(\nabla_{\delta} \mathcal{L}_{\text{ID}}\) and the score-based adversarial conceptual bridge \(S^*\). To ensure dimensional consistency, \(S^*\) (originally computed in the latent score space) is first upsampled and channel-aligned to match the pixel domain of \(\delta\), resulting in \(S^*_{\text{up}}\). Both gradients are then \(\ell_2\)-normalized to balance their relative contributions. The combined gradient is computed as:
\begin{equation}
\text{total\_grad} \leftarrow 
\frac{\nabla_{\delta} \mathcal{L}_{\text{ID}}}{\|\nabla_{\delta} \mathcal{L}_{\text{ID}}\|_2} 
- \frac{S^*_{\text{up}}}{\|S^*_{\text{up}}\|_2}.
\end{equation}
The perturbation is updated via a projected gradient step:
\begin{equation}
\delta \leftarrow \delta - \alpha \cdot \mathrm{sign}(\text{total\_grad}),
\end{equation}
where \(\alpha\) controls the update magnitude. For details of the optimization process, please refer to  Appendix~A.1.

\begin{table*}[htbp]
\centering
\caption{Comparison of identity protection against baseline methods. 
\textbf{ISM} is the average identity similarity (ArcFace, FaceNet, VGGFace); \textbf{FQ} is the FaceQNet quality score. “No Protected” denotes generation w/o defense. Best results are in bold.}
\renewcommand{\arraystretch}{1}
\setlength{\tabcolsep}{1.4mm}
\begin{tabular}{lccccccccccccccc}
\specialrule{1.2pt}{0pt}{0pt} %
\multirow{2}{*}{\textbf{Method}} & \multicolumn{2}{c}{\textbf{Ip-Adapter}} & \multicolumn{2}{c}{\textbf{Ip-Ada + XL}} & \multicolumn{2}{c}{\textbf{Blip-Diffu}} & \multicolumn{2}{c}{\textbf{InfiniteID}} & \multicolumn{2}{c}{\textbf{Photomaker}} & \multicolumn{2}{c}{\textbf{Dreambooth}} & \multicolumn{2}{c}{\textbf{InfiniteYou}} \\
\cmidrule(lr){2-3} \cmidrule(lr){4-5} \cmidrule(lr){6-7} \cmidrule(lr){8-9} \cmidrule(lr){10-11} \cmidrule(lr){12-13} \cmidrule(lr){14-15}
& ISM$\downarrow$ & FQ$\downarrow$ & ISM$\downarrow$ & FQ$\downarrow$ & ISM$\downarrow$ & FQ$\downarrow$ & ISM$\downarrow$ & FQ$\downarrow$ & ISM$\downarrow$ & FQ$\downarrow$ & ISM$\downarrow$ & FQ$\downarrow$ & ISM$\downarrow$ & FQ$\downarrow$ \\
\midrule
\textbf{No Protected} & 0.382 & 0.334 & 0.582 & 0.456 & 0.336 & 0.331 & 0.675 & 0.495 & 0.327 & 0.456 & 0.525 & 0.332 & 0.673 & 0.493 \\
\textbf{Anti-DB \cite{van2023anti}}      & 0.307 & 0.305 & 0.548 & 0.455 & 0.271 & 0.357 & 0.620 & 0.496 & 0.272 & 0.288 & 0.433 & 0.319 & 0.594 & 0.467 \\
\textbf{SimAC \cite{wang2024simac}}        & 0.295 & 0.312 & 0.401 & 0.451 & 0.305 & 0.363 & 0.591 & \textbf{0.484} & 0.248 & 0.296 & 0.390 & 0.343 & 0.571 & 0.459 \\
\textbf{AdvDM \cite{liang2023adversarial}}        & 0.286 & 0.349 & 0.398 & 0.438 & 0.260 & 0.355 & 0.542 & 0.490 & 0.227 & 0.293 & 0.337 & \textbf{0.192} & 0.563 & 0.459 \\
\textbf{ACE \cite{zheng2023understanding}}          & 0.327 & 0.363 & 0.536 & 0.453 & 0.275 & 0.363 & 0.576 & 0.499 & 0.252 & 0.287 & 0.420 & 0.215 & 0.635 & 0.461 \\
\textbf{IDProtector \cite{Song_2025_CVPR}}     & 0.291 & 0.337 & 0.425 & \textbf{0.432} & 0.263 & 0.368 & 0.603 & 0.489 & 0.240 & 0.312 & 0.402 & 0.225 & 0.541 & 0.454 \\
\textbf{IDGuar (Ours)} & \textbf{0.036} & \textbf{0.282} & \textbf{0.126} & 0.450 & \textbf{0.220} & \textbf{0.351} & \textbf{0.444} & 0.494 & \textbf{0.142} & \textbf{0.273} & \textbf{0.336} & 0.220 & \textbf{0.301} & \textbf{0.452} \\
\specialrule{1.2pt}{0pt}{0pt} %
\end{tabular}

\label{tab:performance_comparison}
\end{table*}

\begin{table}[htbp]
\centering
\caption{Comparison of protected image quality (PSNR (dB), SSIM) and time cost (s) across different methods. \textbf{At-DB} denotes \textit{Anti-DreamBooth}, 
\textbf{SAC} denotes \textit{SimAC}, 
\textbf{ADM} denotes \textit{AdvDM}, 
\textbf{ACE} denotes \textit{ACE}, 
\textbf{IDP-P} denotes the PGD-optimized variant of \textit{IDProtector},
\textbf{IDP} denotes \textit{IDProtector}, 
and \textbf{IDGuar} denotes our proposed \textit{IDGuardian}.}
\renewcommand{\arraystretch}{1}
\setlength{\tabcolsep}{0.65mm} 
\begin{tabular}{ccccccccc}
\toprule
 & \textbf{At-DB}& \textbf{SAC}& \textbf{ADM}& \textbf{ACE} &\textbf{IDP-P}&\textbf{IDP}& \textbf{IDGuar}\\
 \midrule
\textbf{PSNR}& 31.63& 32.15& 30.76& 31.61& \textbf{32.49}& 32.10 & 32.19\\
 \textbf{SSIM}& 0.701& 0.772& 0.735& 0.753& 0.840 & 0.805 & \textbf{0.842}\\
\textbf{Time}& 312.1 & 347.3 & 55.3& 90.0 & 4089 & \textbf{0.42} & 47.0\\ 
\bottomrule
\end{tabular}

\label{tab:psnr}
\end{table}

\section{Experiments}

\subsection{Implementation Details and Metrics}
\noindent\textbf{Datasets.} 
We evaluate our method on two widely used face datasets: VGGFace2 \cite{parkhi2015deep} and CelebA-HQ \cite{liu2018large}. Unless specified, primary experiments are conducted on VGGFace2. We exclude images smaller than \(250 \times 250\) pixels, perform alignment and cropping for consistent facial positioning, and select 50 identities, resizing images to \(224 \times 224\). To assess cross-dataset generalization, we further evaluate on CelebA-HQ by applying adversarial perturbations and identity-conditioned generation to 50 high-resolution images (\(512 \times 512\)).

\noindent\textbf{Training Details.} 
We adopt IP-Adapter SDXL Plus Face \cite{ye2023ip} as the surrogate model, built on the pretrained SDXL backbone \cite{podell2023sdxl} with a CLIP-based image encoder to extract identity embeddings for conditioning the UNet. Unless specified, all primary experiments use this SDXL-based surrogate. 
We apply Projected Gradient Descent (PGD) with a fixed text prompt \(y\): ``A photo of a person'', perturbation budget \( \epsilon = 8/255 \), learning rate \( \alpha = 0.005 \), and maximum steps \( N_{\text{iter}} = 200 \). Experiments are conducted on an NVIDIA RTX A6000 GPU (48G). 
To evaluate surrogate-level generalization, we also test a variant based on Stable Diffusion 1.5 (SD~1.5).

\noindent\textbf{Evaluation Metrics.} 
We adopt three categories of evaluation metrics to comprehensively assess identity protection performance: \textit{i) \textbf{Identity Similarity Metrics (ISM)}}: ISM is the average similarity between generated and source images computed from ArcFace, FaceNet, and VGG-Face features, indicating identity removal effectiveness.
\textit{ii)  \textbf{Visual Quality}}: PSNR and SSIM assess the pixel-level fidelity of protected images, reflecting the visual quality after perturbation.
\textit{iii) \textbf{Recognition Utility}}: FaceQNet and MagFace assess the utility of identity-personalized face images for face recognition models.
Details specific to DreamBooth evaluation are provided in Appendix A.2.

\noindent\textbf{Reproducibility Resources.}
For reproducibility, more details on the public resources used in our implementation and evaluation are provided in Appendix~A.3.

\begin{table*}[htbp]
\centering
\caption{Robustness evaluation against common image post-processing operations, including \textbf{JPEG} compression, \textbf{Gaussian blur}, and \textbf{Gaussian noise}, and attacks targeting active defenses such as \textbf{Impress}.}
\renewcommand{\arraystretch}{1}
\setlength{\tabcolsep}{1.4mm}
\begin{tabular}{lccccccccccccccc}
\specialrule{1.2pt}{0pt}{0pt} %
\multirow{2}{*}{\textbf{Method}} & \multicolumn{2}{c}{\textbf{Ip-Adapter}} & \multicolumn{2}{c}{\textbf{Ip-Ada + XL}} & \multicolumn{2}{c}{\textbf{Blip-Diffu}} & \multicolumn{2}{c}{\textbf{InfiniteID}} & \multicolumn{2}{c}{\textbf{Photomaker}} & \multicolumn{2}{c}{\textbf{Dreambooth}} & \multicolumn{2}{c}{\textbf{InfiniteYou}} \\
\cmidrule(lr){2-3} \cmidrule(lr){4-5} \cmidrule(lr){6-7} \cmidrule(lr){8-9}
\cmidrule(lr){10-11} \cmidrule(lr){12-13} \cmidrule(lr){14-15}
& ISM$\downarrow$ & FQ$\downarrow$ & ISM$\downarrow$ & FQ$\downarrow$ & ISM$\downarrow$ & FQ$\downarrow$ & ISM$\downarrow$ & FQ$\downarrow$ & ISM$\downarrow$ & FQ$\downarrow$ & ISM$\downarrow$ & FQ$\downarrow$ & ISM$\downarrow$ & FQ$\downarrow$ \\
\midrule
\textbf{No Protected} 
& 0.382 & 0.334 & 0.582 & 0.456 & 0.336 & 0.331 & 0.675 & 0.495 & 0.327 & 0.456 & 0.525 & 0.332 & 0.673 & 0.493 \\
\textbf{JPEG} 
& 0.141 & \textbf{0.221} & 0.340 & 0.454 & 0.215 & 0.323 & 0.384 & \textbf{0.492} & \textbf{0.135} & 0.274 & 0.331 & 0.254 & 0.322 & 0.454 \\
\textbf{Blur} 
& 0.220 & 0.321 & 0.376 & 0.467 & 0.247 & \textbf{0.290} & \textbf{0.412} & 0.503 & 0.147 & 0.275 & 0.405 & 0.281 & 0.316 & \textbf{0.448} \\
\textbf{Noise} 
& 0.149 & 0.250 & 0.339 & 0.450 & \textbf{0.185} & 0.346 & 0.482 & 0.502 & 0.148 & 0.281 & 0.400 & 0.358 & 0.356 & 0.453 \\
\textbf{Impress} 
& 0.132 & 0.250 & 0.329 & \textbf{0.427} & 0.240 & 0.335 & 0.462 & 0.494 & 0.183 & 0.276 & \textbf{0.174} & 0.248 & \textbf{0.293} & 0.454 \\
\textbf{IDGuar (Ours)} 
& \textbf{0.036} & 0.282 & \textbf{0.126} & 0.450 & 0.220 & 0.351 & 0.444 & 0.494 & 0.142 & \textbf{0.273} & 0.336 & \textbf{0.220} & 0.301 & 0.452 \\
\specialrule{1.2pt}{0pt}{0pt} %
\end{tabular}

\label{tab:Robustness}
\end{table*}

\begin{table*}[htbp]
\centering
\caption{Ablation study of \textbf{identity loss} combinations and \textbf{guidance directions}, where \textbf{G adv} denotes guidance solely toward the adversarial identity distribution and \textbf{G clean} denotes guidance solely away from the clean identity distribution.}
\renewcommand{\arraystretch}{1}
\setlength{\tabcolsep}{1.4mm}
\begin{tabular}{lccccccccccccccc}
\specialrule{1.2pt}{0pt}{0pt} %
\multirow{2}{*}{\textbf{Method}} & \multicolumn{2}{c}{\textbf{Ip-Adapter}} & \multicolumn{2}{c}{\textbf{Ip-Ada + XL}} & \multicolumn{2}{c}{\textbf{Blip-Diffu}} & \multicolumn{2}{c}{\textbf{InfiniteID}} & \multicolumn{2}{c}{\textbf{Photomaker}} & \multicolumn{2}{c}{\textbf{Dreambooth}} & \multicolumn{2}{c}{\textbf{InfiniteYou}} \\
\cmidrule(lr){2-3} \cmidrule(lr){4-5} \cmidrule(lr){6-7} \cmidrule(lr){8-9}
\cmidrule(lr){10-11} \cmidrule(lr){12-13} \cmidrule(lr){14-15}
& ISM$\downarrow$ & FQ$\downarrow$ & ISM$\downarrow$ & FQ$\downarrow$ & ISM$\downarrow$ & FQ$\downarrow$ & ISM$\downarrow$ & FQ$\downarrow$ & ISM$\downarrow$ & FQ$\downarrow$ & ISM$\downarrow$ & FQ$\downarrow$ & ISM$\downarrow$ & FQ$\downarrow$ \\
\midrule
\textbf{No Protected} & 0.382 & 0.334 & 0.582 & 0.456 & 0.336 & 0.331 & 0.675 & 0.495 & 0.327 & 0.456 & 0.525 & 0.332 & 0.673 & 0.493 \\
\midrule
\textbf{CLIP} & 0.243 & 0.313 & 0.217 & 0.473 & 0.287 & 0.368 & 0.577 & 0.514 & 0.264 & 0.301 & 0.504 & 0.331 & 0.531 & 0.473 \\
\textbf{Facenet} & 0.246 & 0.319 & 0.222 & 0.479 & 0.288 & 0.372 & 0.583 & 0.508 & 0.268 & 0.308 & 0.503 & 0.330 & 0.531 & 0.466 \\
\textbf{CLIP + Facenet} & 0.175 & \textbf{0.271} & 0.167 & 0.469 & 0.255 & 0.378 & 0.536 & 0.513 & 0.213 & 0.292 & 0.497 & 0.298 & 0.477 & 0.468 \\
\textbf{IDGuar (Ours)} & \textbf{0.036} & 0.282 & \textbf{0.126} & \textbf{0.450} & \textbf{0.220} & \textbf{0.351} & \textbf{0.444} & \textbf{0.494} & \textbf{0.142} & \textbf{0.273} & \textbf{0.336} & \textbf{0.220} & \textbf{0.301} & \textbf{0.452} \\
\midrule
\textbf{G adv} & 0.044 & 0.290 & 0.144 & 0.460 & 0.285 & 0.357 & 0.468 & \textbf{0.490} & 0.175 & 0.281 & 0.347 & 0.275 & 0.307 & 0.454 \\
\textbf{G clean} & 0.049 & \textbf{0.269} & 0.138 & 0.453 & 0.239 & 0.356 & 0.453 & 0.497 & 0.196 & 0.279 & 0.376 & 0.227 & 0.314 & \textbf{0.452} \\
\textbf{IDGuar (Ours)} & \textbf{0.036} & 0.282 & \textbf{0.126} & \textbf{0.450}& \textbf{0.220} & \textbf{0.351} & \textbf{0.444} & 0.494 & \textbf{0.142} & \textbf{0.273} & \textbf{0.336} & \textbf{0.220} & \textbf{0.301} & \textbf{0.452} \\
\specialrule{1.2pt}{0pt}{0pt} %
\end{tabular}

\label{tab:Ablation}
\end{table*}

\subsection{Overall Performance}

\noindent{\textbf{IDGuardian Enhances Identity Protection Effectiveness.}}
We evaluate IDGuardian against five representative active defense methods across seven mainstream identity-personalization pipelines. These pipelines encompass both training-free methods (IP-Adapter \cite{ye2023ip}, IP-Adapter SDXL Plus Face \cite{ye2023ip}, Blip-Diffusion \cite{li2023blip}, InfiniteID \cite{wu2024infinite}, and PhotoMaker \cite{li2024photomaker}) and training-based methods (DreamBooth \cite{ruiz2023dreambooth} and InfiniteYou \cite{jiang2025infiniteyou}), leveraging diverse generative backbones with Stable Diffusion 1.5, SDXL, and FLUX.1-dev (Transformer-based) for Infinite You.

As summarized in Table~\ref{tab:performance_comparison}, IDGuardian consistently achieves the lowest average identity similarity scores across ArcFace, FaceNet, and VGGFace metrics, alongside substantial reductions in image quality scores measured by FaceQNet. Detailed per-metric results and comprehensive statistical analysis are provided in Appendix~B.1.
Qualitative examples in Figure~\ref{fig:vis} further confirm that IDGuardian effectively suppresses identity cues across a wide range of personalization scenarios, outperforming competing defense methods. These findings highlight the generalizability of IDGuardian in protecting identity across diverse generative models and personalization pipelines.

\begin{figure}[htbp]
    \centering
    \includegraphics[width=0.47\textwidth]{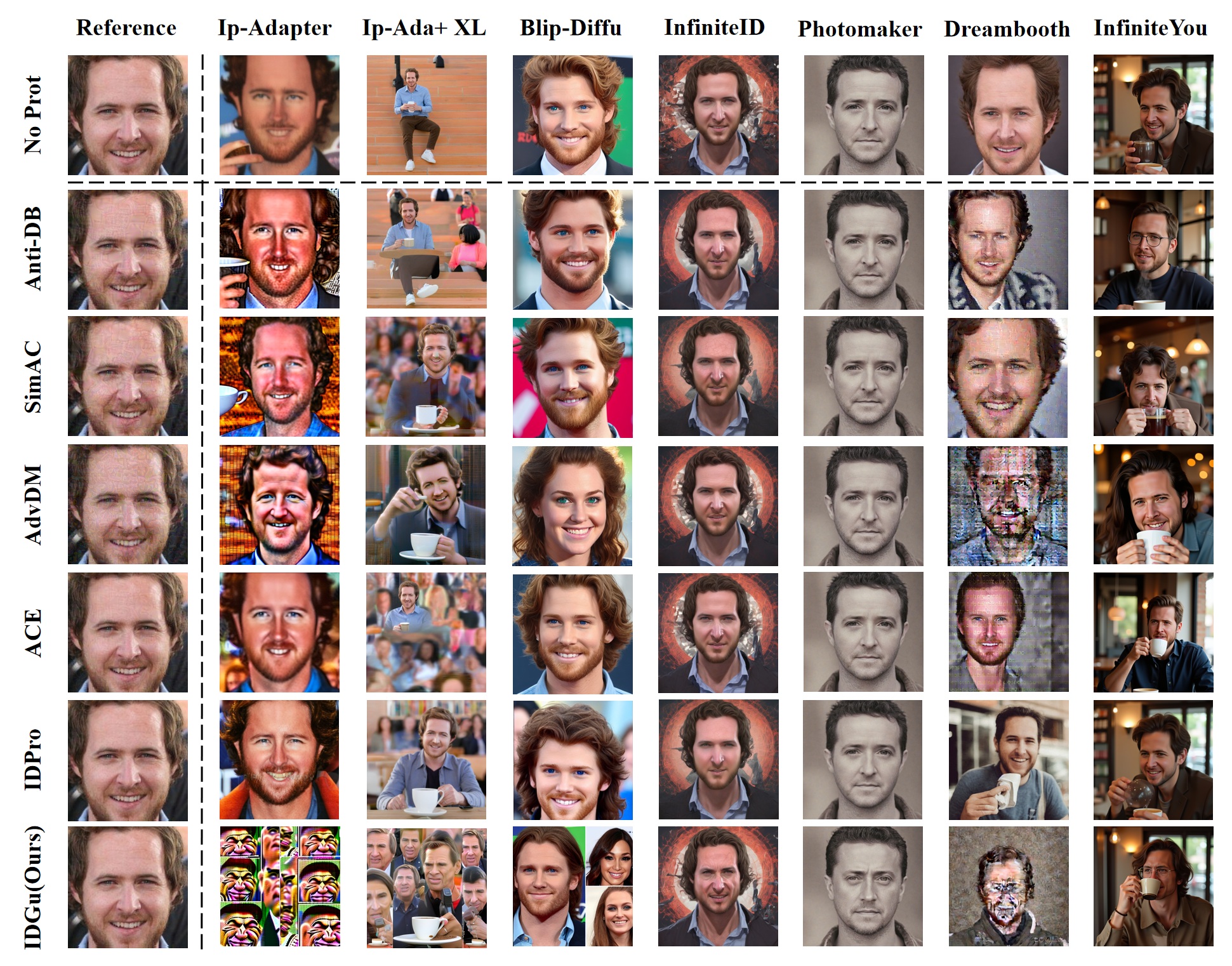}
    \caption{Visualization of \textbf{identity suppression effectiveness} for IDGuardian and comparison methods across different personalization settings.}
    \label{fig:vis}
\end{figure}

\noindent\textbf{IDGuardian Preserves Perceptual Quality While Suppressing Identity.}
Table~\ref{tab:psnr} reports PSNR, SSIM, and time cost scores evaluating visual fidelity and computational efficiency of IDGuardian-processed images. Results show that, compared with other defense methods, IDGuardian more effectively suppresses identity cues while better preserving perceptual quality and maintaining competitive time efficiency.  
The low time cost of IDGuardian arises from its training-free inference design. By contrast, IDP requires training a noise generator that can generalize to most images, incurring substantial computational resources. This is reflected in IDP-P, where performing PGD optimization for each image clearly results in much higher time cost.

\noindent\textbf{IDGuardian Is Robust To Post-Processing and Adaptive Threats.}
We assess the robustness of IDGuardian against common image degradations—including JPEG compression, Gaussian blur, and Gaussian noise—as well as against adaptive threats such as Impress~\cite{cao2023impress}, a recent method designed to bypass active defenses. As shown in Table~\ref{tab:Robustness}, IDGuardian consistently maintains effective identity suppression under these challenging conditions.
Interestingly, for some baselines, mild post-processing such as JPEG compression may slightly improve defense performance. While compression generally disrupts adversarial perturbations, in some cases it also perturbs the model's recognition of the original image's identity features, which can inadvertently make identity suppression appear stronger. 
Qualitative examples in Figure~8 of the Appendix further support these results. Please see Appendix~B.3 and ~B.4 for details.

\noindent\textbf{IDGuardian Generalizes to Commercial APIs.}  
We evaluate IDGuardian on commercial face personalization APIs (Seedream 4.0, Qwen-Image, Kling-AI, Ihuiwa). As shown in Appendix~D.1, it consistently suppresses identity similarity, demonstrating robustness under black-box conditions.

\noindent\textbf{IDGuardian Generalizes Across Surrogate Models.}
To assess the generalization of IDGuardian, we replace the default SDXL-based IP-Adapter with a variant based on Stable Diffusion 1.5. As shown in Table~2 and Figure~6 of Appendix, IDGuardian consistently achieves strong identity suppression across both generative backbones, demonstrating robustness to architectural variations in the surrogate model. Further details are provided in Appendix~D.2.

\noindent\textbf{IDGuardian Generalizes Across Datasets.}
Beyond VGGFace2, we further evaluate IDGuardian on the CelebA-HQ dataset to assess generalization across data distributions.  As shown in Table~9, Table~10, and Figure~10 of Appendix, IDGuardian consistently suppresses identity similarity across various metrics. 
Consistent results on the LFW dataset using the identity similarity metric are further reported in Table~11 of the Appendix.
This performance holds true across different image resolutions and demographics, reinforcing the robustness of our method in diverse real-world scenarios. For details, please refer to Appendix~D.3.

\noindent\textbf{Human and GPT-Based Evaluation of Identity Protection Efficacy.}
To further validate the effectiveness of identity protection, we conduct both a user study and a GPT-based evaluation using a large multimodal model to assess perceived identity similarity. Results from both evaluations consistently indicate that IDGuardian significantly reduces identity similarity compared to baseline methods. For details on the evaluation protocols and results, please refer to Appendix~E.1, ~E.2 and Figure~7 of Appendix.

\subsection{Ablation Study}

\noindent\textbf{Roles of Individual Components.}
We conduct ablations on three identity supervision components: i) CLIP identity loss alone, ii) combined CLIP and FaceNet losses, and iii) CLIP + FaceNet losses with additional Score-Based Adversarial Conceptual Bridge. As shown in Table~\ref{tab:Ablation}, the dual-encoder setup improves identity suppression by leveraging both semantic and low-level features. Incorporating adversarial conceptual bridge further enhances suppression effectiveness significantly, demonstrating the key role of guided optimization. More details are provided in Appendix~B.2.

\noindent\textbf{Adversarial Guidance Direction.}
We study how the direction of adversarial score guidance influences optimization. We compare i)  gradient toward the adversarial identity distribution ($\nabla_{\mathbf{x}_t} \log p(\mathbf{y}_{\text{adv}} \mid \mathbf{x}_t)$), and ii) gradient away from the clean identity distribution ($\nabla_{\mathbf{x}_t} \log p(\mathbf{y}_{\text{clean}} \mid \mathbf{x}_t)$). Both are one-sided and often unstable due to distributional ambiguity, while our direction explicitly models the shift and achieves more stable identity suppression. Results are shown in Table~\ref{tab:Ablation}. Qualitative visualizations are provided in Figure~9 of the Appendix. See Appendix~B.2 for details.

\noindent\textbf{Perturbation Budget and Learning Rate.}
We perform a grid search over perturbation magnitude $\epsilon$ and step size $\alpha$ to explore their effects on identity suppression and perceptual quality. As shown in Appendix Figure 3, the combination $\alpha = 0.005$ and $\epsilon = 8/255$ achieves the best balance, effectively reducing identity similarity while preserving visual fidelity. Smaller perturbations yield insufficient suppression, while larger ones introduce artifacts and degrade quality. Please refer to Appendix~C.2 for more details.

\noindent\textbf{Number Of Optimization Steps.}
We analyze convergence by tracking identity loss over 1000 iterations. As shown in Figure~4 of the Appendix, loss decreases rapidly in the first 100 steps and stabilizes after 200, with only marginal gains afterward. Thus, we set optimization steps to 200 for all experiments. Further details are in Appendix~C.3.

\section{Conclusion}


We propose \textbf{IDGuardian}, an adversarial defense framework for proactive identity protection in diffusion- and flux-based personalized image generation. By learning imperceptible perturbations, it prevents generated outputs from preserving the original identity under similarity-based evaluation. IDGuardian integrates two complementary strategies: \textbf{i) Cross-Encoder Identity Field Confounding}, which reduces identity similarity across multiple encoders, and \textbf{ii) Guidance-Flow Identity Deflection}, which steers the diffusion trajectory away from the target identity manifold. It remains imperceptible while effectively suppressing identity information, and is robust to common post-processing operations and adaptive removal attacks such as Impress. Experiments show that IDGuardian reduces identity consistency in generated outputs, mitigating identity leakage in personalized image synthesis. 


\clearpage
\section*{Acknowledgments}
This work was supported in part by the National Natural Science Foundation of China under Grants 62472231, 62172233, and 62502215; the Jiangsu Provincial Science and Technology Major Project under Grant BG2024042; the Natural Science Foundation of Jiangsu under Grant BK20250735; and the Postgraduate Research \& Practice Innovation Program of Jiangsu Province under Grant SJCX25\_0523.

{
    \small
    \bibliographystyle{ieeenat_fullname}
    \bibliography{main}
}

\appendix
 \clearpage
\setcounter{page}{1}
\maketitlesupplementary

\section{Methodological and Evaluation Details}

While the main paper introduces the two-stage defense principle of \textbf{IDGuardian}, this section provides the complete algorithmic details of its optimization process. Specifically, IDGuardian jointly optimizes adversarial perturbations through two complementary strategies—\textbf{Cross-Encoder Identity Field Confounding } and \textbf{Guidance-Flow Identity Deflection}—which correspond to the disruption of identity extraction and injection, respectively. 
For clarity and reproducibility, we detail the optimization objectives, loss composition, and implementation settings that were omitted for brevity in the main text.

\subsection{Optimization Process of IDGuardian}
\label{app:A.1}


Our defense operates in the latent denoising space of the pretrained SDXL backbone within the IP-Adapter SDXL Plus Face pipeline, and optimizes an adversarial perturbation \(\delta\) on the input image \(x\), such that the final protected image \(x_{\text{adv}} = x + \delta\) preserves perceptual quality while significantly reducing identity consistency in downstream personalized generation.

The optimization is carried out iteratively over \(N_{\text{iter}}\) steps. At each step:
\begin{itemize}
    \item The input clean image is encoded into the latent space using the VAE encoder.
    \item A random diffusion timestep \(t\) is sampled, and corresponding noise is added to simulate a forward noisy latent \(x_t\).
    \item Identity embeddings are computed from the clean and perturbed versions of the input, and serve as conditioning inputs for the UNet during denoising.
    \item Two complementary gradients are computed: i) the identity loss gradient \(\nabla_{\delta} \mathcal{L}_{\text{ID}}\) based on FaceNet and CLIP similarity; ii) a distributional score gradient \(S^*\), which measures the semantic divergence between adversarial and clean identity trajectories in diffusion space.
\item To match the spatial resolution and RGB channels of the perturbation, $S^*$ is upsampled via bilinear interpolation (PyTorch: \textit{F.interpolate(..., mode='bilinear', align\_corners=False)}) and only the first three channels (RGB) are retained to align with the perturbation tensor.
    \item These gradients are normalized and combined to update \(\delta\) via a projected gradient descent step, bounded by the perturbation constraint \(\|\delta\|_\infty < \epsilon\).
\end{itemize}

Algorithm~\ref{alg:IDGuardian} outlines the full optimization pipeline. This instantiation reflects a synergy between semantic feature-space objectives and internal generative guidance dynamics, ensuring that the optimization is both perceptually aware and diffusion-aligned.

\begin{algorithm}
\small
\setlength{\baselineskip}{12.3pt}
\caption{IDGuardian}
\begin{algorithmic}[1]
\STATE \textbf{Input:} original image \(x\), perturbation budget \(\epsilon\), learning rate \(\alpha\), text prompt \(y\), max steps \(N_{\text{iter}}\)
\STATE \textbf{Output:} optimized adversarial image \(x_{\text{adv}}\)
\STATE Load pretrained \textbf{IP-Adapter SDXL Plus Face} pipeline components: VAE Encoder, UNet, and CLIP image encoder
\STATE Encode image into latent space: \(x_0 \leftarrow \text{VAE\_Encoder}(x)\)
\STATE Initialize perturbation \(\delta \leftarrow 0\)
\FOR{\(i = 1\) to \(N_{\text{iter}}\)}
    \STATE Sample random timestep \(t\), corresponding noise \(\epsilon_t\)
    \STATE Compute noisy latent: \(x_t \leftarrow \text{AddNoise}(x_0, \epsilon_t, t)\)
    \STATE \(x_{\text{adv}} \leftarrow \text{clip}(x + \delta, 0, 1)\)
    \STATE Compute embeddings: \(\text{CLIP}(x), \text{CLIP}(x_{\text{adv}})\)
    \STATE Predict denoising outputs:
    \STATE \quad \(\hat{\epsilon}_{\text{clean}} \leftarrow \text{UNet}(x_t, t, \text{CLIP}(x), y)\)  
    \STATE \quad \(\hat{\epsilon}_{\text{adv}} \leftarrow \text{UNet}(x_t, t, \text{CLIP}(x_{\text{adv}}), y)\)
    \STATE Compute score gradient:
    \STATE \quad \(S^* \leftarrow -\frac{1}{\sqrt{1 - \bar{\alpha}_t}} \cdot (\hat{\epsilon}_{\text{adv}} - \hat{\epsilon}_{\text{clean}})\)
     \STATE Upsample \(S^*\) to image resolution using bilinear interpolation and align channels with RGB:
    \STATE \quad \(S^*_{\text{up}} \leftarrow \text{BilinearUp}(S^*, \text{size}=x.\text{shape})[:, :3, :, :]\)
    \STATE Compute identity loss gradient: \(\nabla_{\delta} \mathcal{L}_{\text{ID}}\)
    \STATE Combine gradients: 
    \STATE \quad \(\text{total\_grad} \leftarrow \text{Normalize}(\nabla_{\delta} \mathcal{L}_{\text{ID}}) - \text{Normalize}(S^*_{\text{up}})\)
    \STATE Update perturbation: \(\delta \leftarrow \delta - \alpha \cdot \text{sign}(\text{total\_grad})\)
    \STATE Clip: \(\delta \leftarrow \text{clip}(\delta, -\epsilon, \epsilon)\)
\ENDFOR
\STATE \textbf{Return} \(x_{\text{adv}} = x + \delta\)
\end{algorithmic}
\label{alg:IDGuardian}
\end{algorithm}

\subsection{DreamBooth Fine-Tuning and Similarity Evaluation}
\label{app:A.2}

For DreamBooth and other training-based methods, which typically require multiple images per identity for fine-tuning, we provide four distinct images per identity. For each generated image of that identity, we compute its similarity to each of the four original images, yielding four similarity scores. We then average these four scores to obtain a single similarity metric for that generated image. This procedure is repeated for all generated images of the identity, and the final reported similarity is obtained by averaging across all generated images. This ensures that the evaluation reflects the model's performance using all available reference images rather than relying on a single image.

\subsection{Reproducibility Resources}
\label{app:A.3}

For reproducibility, we list the main public resources used in this work:
\begin{itemize}
    \item \textbf{FaceNet implementation:} We adopt the official FaceNet implementation from the GitHub repository \url{davidsandberg/facenet}, using the \textit{Inception-ResNet-v1} checkpoint with identifier \texttt{20180402-114759} in our loss computation.
    \item \textbf{Evaluation toolkit:} We use the public DeepFace toolkit from the GitHub repository \url{serengil/deepface} for identity similarity evaluation.
    \item \textbf{Datasets:} The VGGFace2 dataset is accessed from the HuggingFace repository \url{datasets/ProgramComputer/VGGFace2}. The CelebA-HQ dataset is obtained from the GitHub repository \url{IIGROUP/MM-CelebA-HQ-Dataset}.
\end{itemize}


\begin{table*}[htbp]
\centering
\caption{Robustness evaluation under different distortion levels. The results are reported on Ip-Adapter Plus XL as an example.}
\renewcommand{\arraystretch}{1.36}
\setlength{\tabcolsep}{2.3mm}
\begin{tabular}{lccccccccccccccccccc}
\specialrule{1.0pt}{0pt}{0pt} 
 \multicolumn{3}{c}{\textbf{No Protected}} & \multicolumn{3}{c}{\textbf{JPEG (85)}} & \multicolumn{3}{c}{\textbf{Blur (5 × 5, 1.5)}} & \multicolumn{3}{c}{\textbf{Noise (100)}}  \\
\cmidrule(lr){1-3} \cmidrule(lr){4-6} \cmidrule(lr){7-9} \cmidrule(lr){10-12}  
\textbf{Arc $\downarrow$} & \textbf{Face $\downarrow$} & \textbf{VGG $\downarrow$} & \textbf{Arc $\downarrow$} & \textbf{Face $\downarrow$} & \textbf{VGG $\downarrow$} & \textbf{Arc $\downarrow$} & \textbf{Face $\downarrow$} & \textbf{VGG $\downarrow$} & \textbf{Arc $\downarrow$} & \textbf{Face $\downarrow$} & \textbf{VGG $\downarrow$} \\
\specialrule{0.8pt}{0pt}{0pt}
0.4985 & 0.6899 & 0.5573 & 0.3209 & 0.3983 & 0.3006 & 0.3317 & 0.4443 & 0.3520 & 0.3078 & 0.3970 & 0.3113 \\
\specialrule{1.0pt}{0pt}{0pt}
 \multicolumn{3}{c}{\textbf{No Protected}} & \multicolumn{3}{c}{\textbf{JPEG (70)}} & \multicolumn{3}{c}{\textbf{Blur (7 × 7, 1.5)}} & \multicolumn{3}{c}{\textbf{Noise (400)}}  \\
\cmidrule(lr){1-3} \cmidrule(lr){4-6} \cmidrule(lr){7-9} \cmidrule(lr){10-12}  
\textbf{Arc $\downarrow$} & \textbf{Face $\downarrow$} & \textbf{VGG $\downarrow$} & \textbf{Arc $\downarrow$} & \textbf{Face $\downarrow$} & \textbf{VGG $\downarrow$} & \textbf{Arc $\downarrow$} & \textbf{Face $\downarrow$} & \textbf{VGG $\downarrow$} & \textbf{Arc $\downarrow$} & \textbf{Face $\downarrow$} & \textbf{VGG $\downarrow$} \\
\specialrule{0.8pt}{0pt}{0pt}
0.4985 & 0.6899 & 0.5573 & 0.3520 & 0.4863 & 0.3730 & 0.3303 & 0.4473 & 0.3338 & 0.3955 & 0.4778 & 0.3746 \\
\specialrule{1.0pt}{0pt}{0pt}
 \multicolumn{3}{c}{\textbf{No Protected}} & \multicolumn{3}{c}{\textbf{JPEG (50)}} & \multicolumn{3}{c}{\textbf{Blur (11 × 11, 1.5)}} & \multicolumn{3}{c}{\textbf{Noise (900)}}  \\
\cmidrule(lr){1-3} \cmidrule(lr){4-6} \cmidrule(lr){7-9} \cmidrule(lr){10-12}  
\textbf{Arc $\downarrow$} & \textbf{Face $\downarrow$} & \textbf{VGG $\downarrow$} & \textbf{Arc $\downarrow$} & \textbf{Face $\downarrow$} & \textbf{VGG $\downarrow$} & \textbf{Arc $\downarrow$} & \textbf{Face $\downarrow$} & \textbf{VGG $\downarrow$} & \textbf{Arc $\downarrow$} & \textbf{Face $\downarrow$} & \textbf{VGG $\downarrow$} \\
\specialrule{0.8pt}{0pt}{0pt}
0.4985 & 0.6899 & 0.5573 & 0.3426 & 0.4606 & 0.3382 & 0.3416 & 0.4728 & 0.3632 & 0.3094 & 0.4495	& 0.3604 \\
\specialrule{1.0pt}{0pt}{0pt}
\end{tabular}

\label{tab:RobustnessEnhanced}
\end{table*}

\begin{figure}[htbp]
    \centering
    \includegraphics[width=0.47\textwidth]{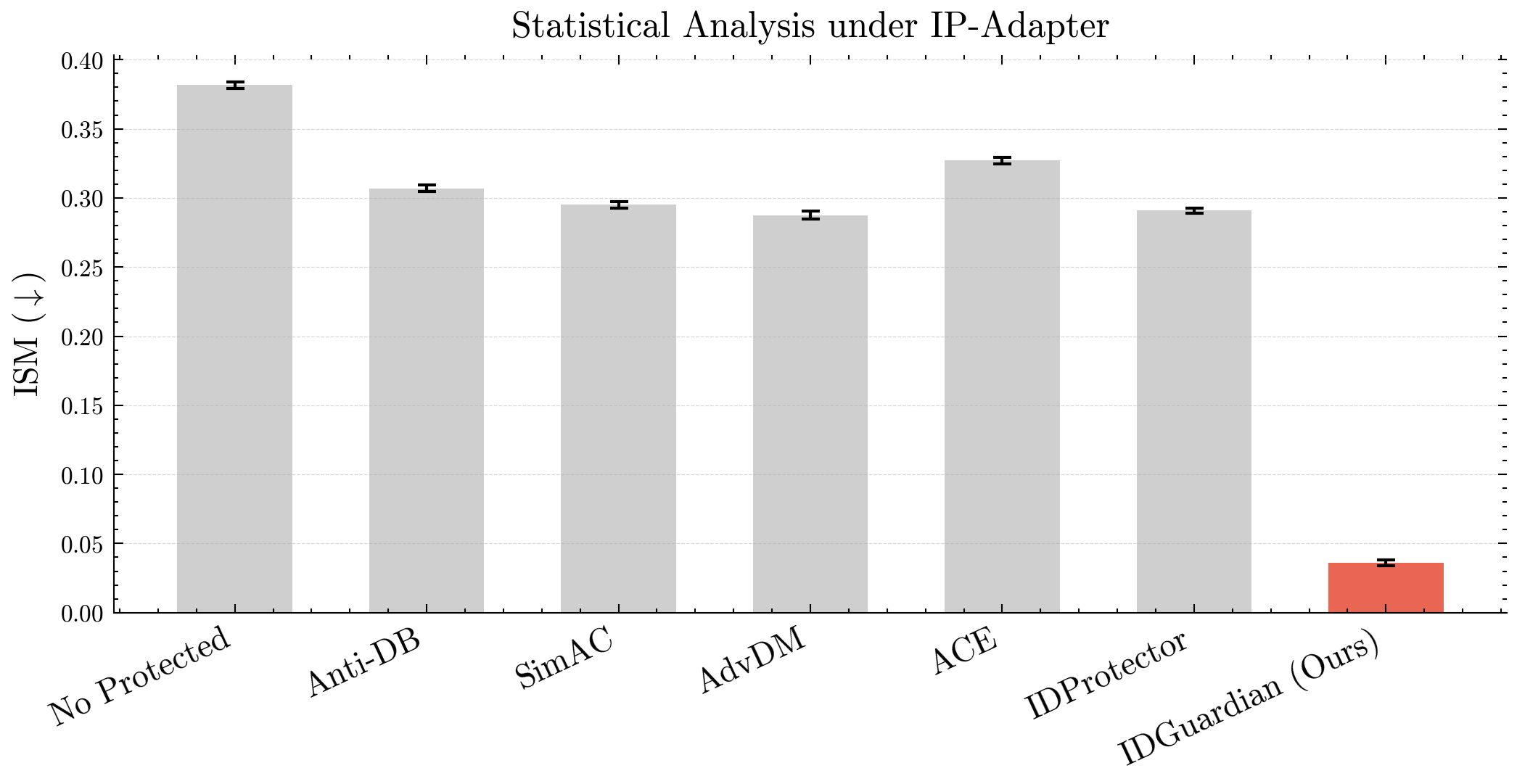}
\caption{Statistical analysis of identity similarity (ISM) under IP-Adapter. Error bars denote the variability across repeated runs.}
    \label{fig:statistical}
\end{figure}

\section{Detailed Quantitative Results}

\subsection{Comparative Evaluation with Baselines}
\label{app:B.1}

In the main paper, we summarized the performance of IDGuardian against baseline methods using average identity similarity scores and a representative perceptual quality metric.  
Here, we provide comprehensive quantitative results for all individual identity similarity metrics (ArcFace, FaceNet, VGGFace) and perceptual quality metrics (FaceQNet, MagFace) across the tested methods.  
Table~\ref{tab:com performance_comparison} and Table~\ref{tab:com performance_comparison1} present detailed comparisons highlighting the consistent superiority of IDGuardian in identity suppression and its effectiveness in preserving perceptual quality.

We further include statistical analysis results, demonstrating that the observed performance gains remain stable and consistent across repeated runs. Taking IP-Adapter as an example, as shown in Figure~\ref{fig:statistical}, IDGuardian exhibits consistently lower identity similarity with small variance compared to baseline methods.

\begin{figure}[htbp]
    \centering
    \includegraphics[width=0.47\textwidth]{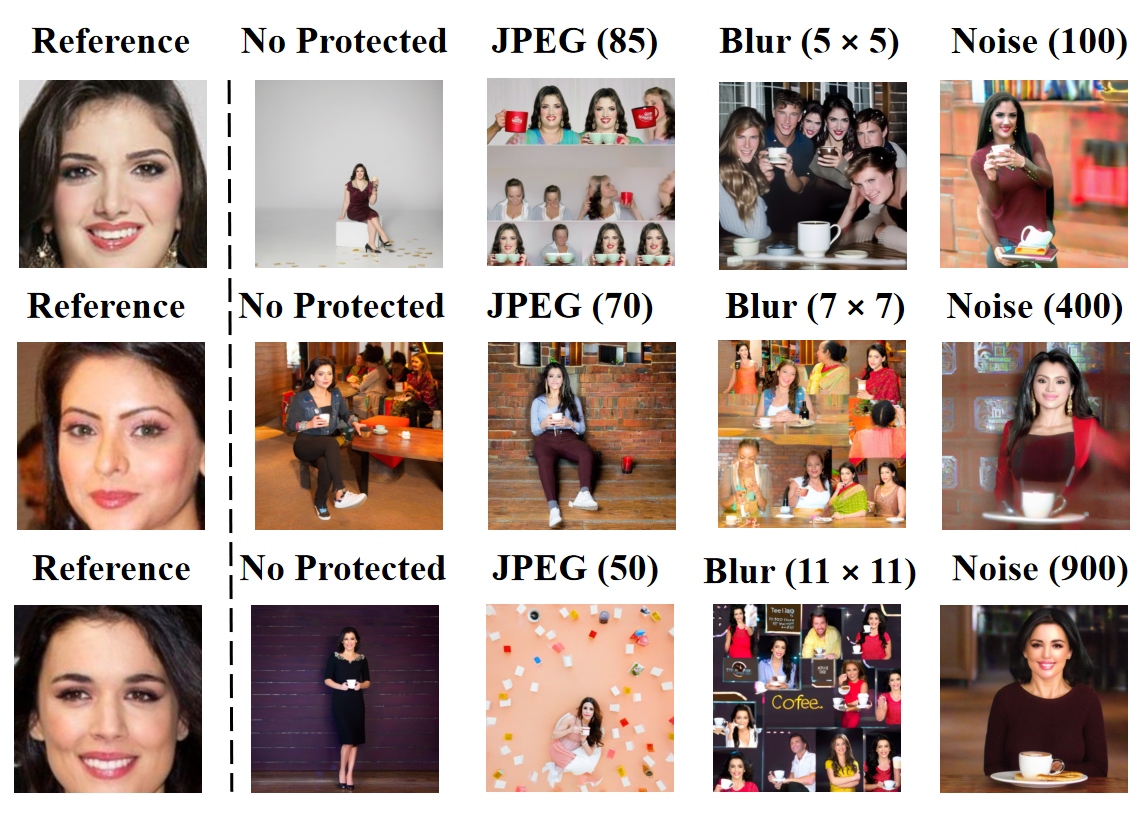}
    \caption{Visual results under three levels of JPEG compression, Gaussian blur, and Gaussian noise, showing that our protection mechanism remains effective even under stronger distortions.}
    \label{fig:RobustnessVis}
\end{figure}

\subsection{Ablation Study on Identity Similarity and Perceptual Quality Metrics}
\label{app:B.2}

We further expand the ablation study presented in the main paper by showing the full set of identity similarity metrics and perceptual quality metrics across various ablation configurations.  
Tables~\ref{tab:Ablation performance_comparison} and~\ref{tab:Ablation performance_comparison1} provide detailed insights into the contributions of different identity encoder combinations and adversarial conceptual bridge.  
These results validate the complementary roles of identity supervision and conceptual bridge in enhancing identity suppression, where visual degradation is part of the intended defense effect.

\subsection{Robustness Evaluation Settings.}
\label{app:B.3}

To evaluate the robustness of our protection mechanism, we applied three types of distortions to the protected images: JPEG compression, Gaussian blur, and Gaussian noise. Specifically, we used JPEG compression with a quality factor of 85, Gaussian blur with a kernel size of $5\times5$ and $\sigma=1.5$, and noise perturbation with a variance of 100. These parameters were selected to simulate common post-processing or transmission degradations while maintaining perceptual plausibility. As illustrated in Figure~5 of the main text, the protection performance under these distortions remains comparable to that on undistorted inputs, demonstrating that our method exhibits strong robustness against typical image degradations.
To further evaluate the robustness, we conducted additional experiments with stronger distortions. Specifically, we tested two additional levels of perturbations: JPEG compression with quality factors of 70 and 50, Gaussian blur with kernel sizes of $7\times7$ and $11\times11$ (both with $\sigma=1.5$), and Gaussian noise with variances of 400 and 900. Table~\ref{tab:RobustnessEnhanced} summarizes the results, demonstrating that our method maintains protection performance even under these more severe image degradations. As illustrated in Figure~\ref{fig:RobustnessVis}, we also provide visual comparisons under three levels of post-processing strength, showing that the perceptual quality of the protected images remains largely preserved despite the increased distortions.

\begin{figure*}[htbp]
    \centering
    \includegraphics[width=1.0\textwidth]{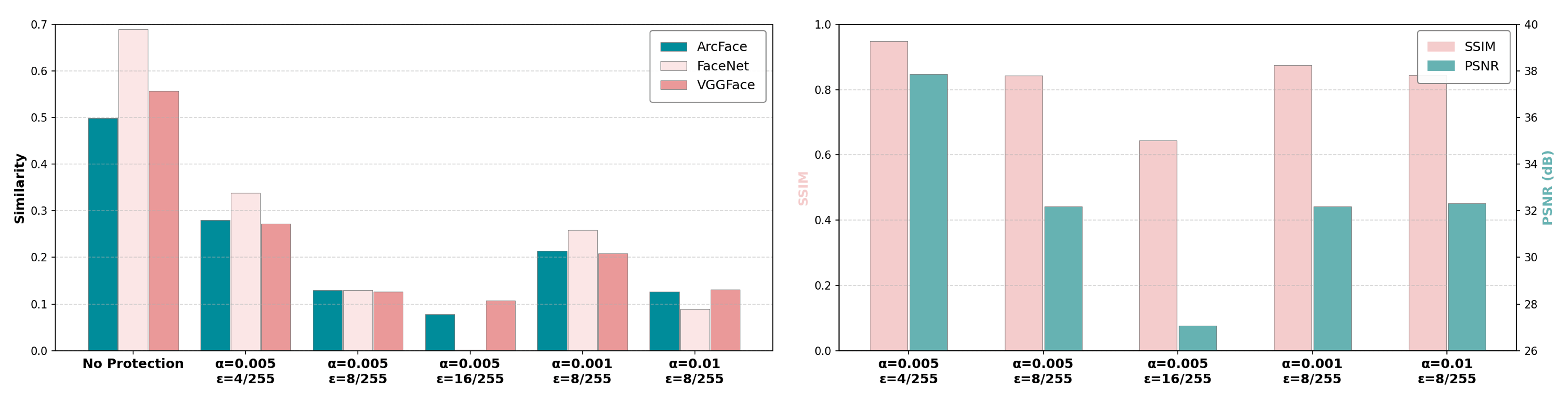}
    \caption{Grid search results for perturbation budget \(\epsilon\) and learning rate \(\alpha\). The configuration \(\alpha=0.005\), \(\epsilon=8/255\) achieves the best trade-off between identity suppression (measured by ArcFace, FaceNet, and VGGFace similarity) and perceptual quality (SSIM and PSNR).}
    \label{fig:PGD Ablation}
\end{figure*}

\subsection{Robustness Evaluation on Identity Similarity and Perceptual Quality Metrics}
\label{app:B.4}

Robustness experiments are conducted under both common image degradations (JPEG compression, Gaussian blur, and additive noise) and adaptive defense-aware attacks such as Impress, which is explicitly designed to circumvent identity protection mechanisms. As shown in Tables~\ref{tab:Robustness performance_comparison} and~\ref{tab:Robustness performance_comparison1}, IDGuardian maintains low identity similarity scores across all perturbation types, while perceptual quality metrics naturally decline under heavy input distortion. These results demonstrate that our method retains strong protection capability even under adverse conditions.
Qualitative examples in Figure~\ref{fig:vis2} further support these findings by illustrating that identity-relevant features remain suppressed even under various post-processing transformations. These detailed robustness results complement the summary metrics presented in the main paper and underscore IDGuardian’s practical effectiveness in real-world usage scenarios.

\section{Additional Ablation Study}

\subsection{Visualization of the Main Ablation Results}
\label{app:C.1}

To further validate the effectiveness of our proposed identity guidance strategy and loss function design, we provide an additional visualization of the ablation results in Figure~\ref{fig:vis4}. The figure clearly demonstrates the influence of different identity loss combinations and guidance directions on the overall performance. Consistent with the quantitative results in Table 4 of the main text, IDGuardian achieves superior identity protection while maintaining high visual quality.

\subsection{Perturbation Budget and Learning Rate}
\label{app:C.2}

We perform a grid search over the perturbation magnitude $\epsilon$ and learning rate $\alpha$ to determine an optimal trade-off between identity consistency suppression and perceptual quality. Specifically, we evaluate the following five configurations:

\begin{itemize}
    \item $\alpha=0.005$, $\epsilon=4/255$
    \item $\alpha=0.005$, $\epsilon=8/255$
    \item $\alpha=0.005$, $\epsilon=16/255$
    \item $\alpha=0.001$, $\epsilon=8/255$
    \item $\alpha=0.01$, $\epsilon=8/255$
\end{itemize}

As depicted in Figure~\ref{fig:PGD Ablation}, the configuration $\alpha=0.005$, $\epsilon=8/255$ achieves the best balance:
\begin{itemize}
    \item It consistently reduces identity consistency below 0.15 across ArcFace, FaceNet, and VGGFace.
    \item It maintains high perceptual quality, with SSIM exceeding 0.8419 and PSNR surpassing 32.186 dB.
\end{itemize}

Smaller perturbations (e.g., $\epsilon=4/255$) yield insufficient identity suppression, while larger ones (e.g., $\epsilon=16/255$) compromise visual fidelity. Similarly, extremely small learning rates ($\alpha=0.001$) lead to under-optimization, whereas large values ($\alpha=0.01$) introduce instability and visual artifacts. These observations underscore the importance of jointly tuning the perturbation magnitude and step size, as a delicate balance is required to ensure both identity confusion and natural-looking outputs.

\begin{table*}[htbp]
\centering
\caption{Identity Suppression Performance Across Surrogate Generative Models. The metrics Arc, Face, and VGG correspond to identity similarity scores computed by the face recognition models ArcFace, FaceNet, and VGGFace, respectively.}
\renewcommand{\arraystretch}{1.40}
\setlength{\tabcolsep}{1.1mm}
\begin{tabular}{lccccccccccccccccccc}
\specialrule{1.0pt}{0pt}{0pt} 
 \multirow{2}{*}{\textbf{Method}} & \multicolumn{3}{c}{\textbf{\textbf{\textbf{Ip-Adapter}}
}} & \multicolumn{3}{c}{\textbf{\textbf{\textbf{Ip-Adapter Plus XL}}}} & \multicolumn{3}{c}{\textbf{\textbf{\textbf{Blip-Diffusion}}}} & \multicolumn{3}{c}{\textbf{\textbf{\textbf{InfiniteID}}}}  \\
\cmidrule(lr){2-4} \cmidrule(lr){5-7} \cmidrule(lr){8-10} \cmidrule(lr){11-13}  
&\textbf{Arc $\downarrow$} & \textbf{Face $\downarrow$} & \textbf{VGG $\downarrow$}&\textbf{Arc $\downarrow$} & \textbf{Face $\downarrow$} & \textbf{VGG $\downarrow$} &\textbf{Arc $\downarrow$} & \textbf{Face $\downarrow$} & \textbf{VGG $\downarrow$}&\textbf{Arc $\downarrow$} & \textbf{Face $\downarrow$} & \textbf{VGG $\downarrow$} \\
\specialrule{0.8pt}{0pt}{0pt}
\textbf{No Protected}  & 0.3369  & 0.4774 & 0.3308 &0.4985 & 0.6899 & 0.5573 &0.3305 & 0.3918 & 0.2848 & 0.6345 & 0.7627 & 0.6286   \\
\textbf{IDGuardian-SD1.5} & \textbf{0.0425 }& \textbf{0.0119} & \textbf{0.0650} &\textbf{0.1589 }& \textbf{0.1413} & \textbf{0.1403} & \textbf{0.2670} & \textbf{0.2550 }& \textbf{0.2204}  & \textbf{0.4338} & \textbf{0.4989 }& \textbf{0.4113}\\
\specialrule{1.0pt}{0pt}{0pt}
 \multirow{2}{*}{\textbf{Method}}  & \multicolumn{3}{c}{\textbf{\textbf{\textbf{Photomaker}}}} & \multicolumn{3}{c}{\textbf{\textbf{\textbf{Dreambooth}}}} & \multicolumn{3}{c}{\textbf{\textbf{\textbf{InfiniteYou}}}} & \multicolumn{3}{c}{\textbf{\textbf{\textbf{Mean}}}}  \\
\cmidrule(lr){2-4} \cmidrule(lr){5-7} \cmidrule(lr){8-10} \cmidrule(lr){11-13} 
& \textbf{Arc $\downarrow$} & \textbf{Face $\downarrow$} & \textbf{VGG $\downarrow$}&\textbf{Arc $\downarrow$} & \textbf{Face $\downarrow$} & \textbf{VGG $\downarrow$} &\textbf{Arc $\downarrow$} & \textbf{Face $\downarrow$} & \textbf{VGG $\downarrow$}&\textbf{Arc $\downarrow$} & \textbf{Face $\downarrow$} & \textbf{VGG $\downarrow$} \\
\specialrule{0.8pt}{0pt}{0pt}
\textbf{No Protected}  &0.3488 & 0.3513 & 0.2806 & 0.4681 & 0.6580 & 0.4477  & 0.6788 & 0.7554 & 0.5847 & 0.4708 & 0.5838 & 0.4450\\
\textbf{IDGuardian-SD1.5}  & \textbf{0.1353} &\textbf{0.1556} & \textbf{0.2233} &\textbf{0.2907 }&\textbf{0.3940} &\textbf{0.3455 }&\textbf{0.3328} &\textbf{0.3573} & \textbf{0.2809} & \textbf{0.2373} & \textbf{0.2591 }& \textbf{0.2409}\\
\specialrule{1.0pt}{0pt}{0pt}
\end{tabular}

\label{tab:surrogate_model_identity}
\end{table*}

\subsection{Number of Optimization Steps}
\label{app:C.3}

\begin{figure}[htbp]
    \centering
    \includegraphics[width=0.47\textwidth]{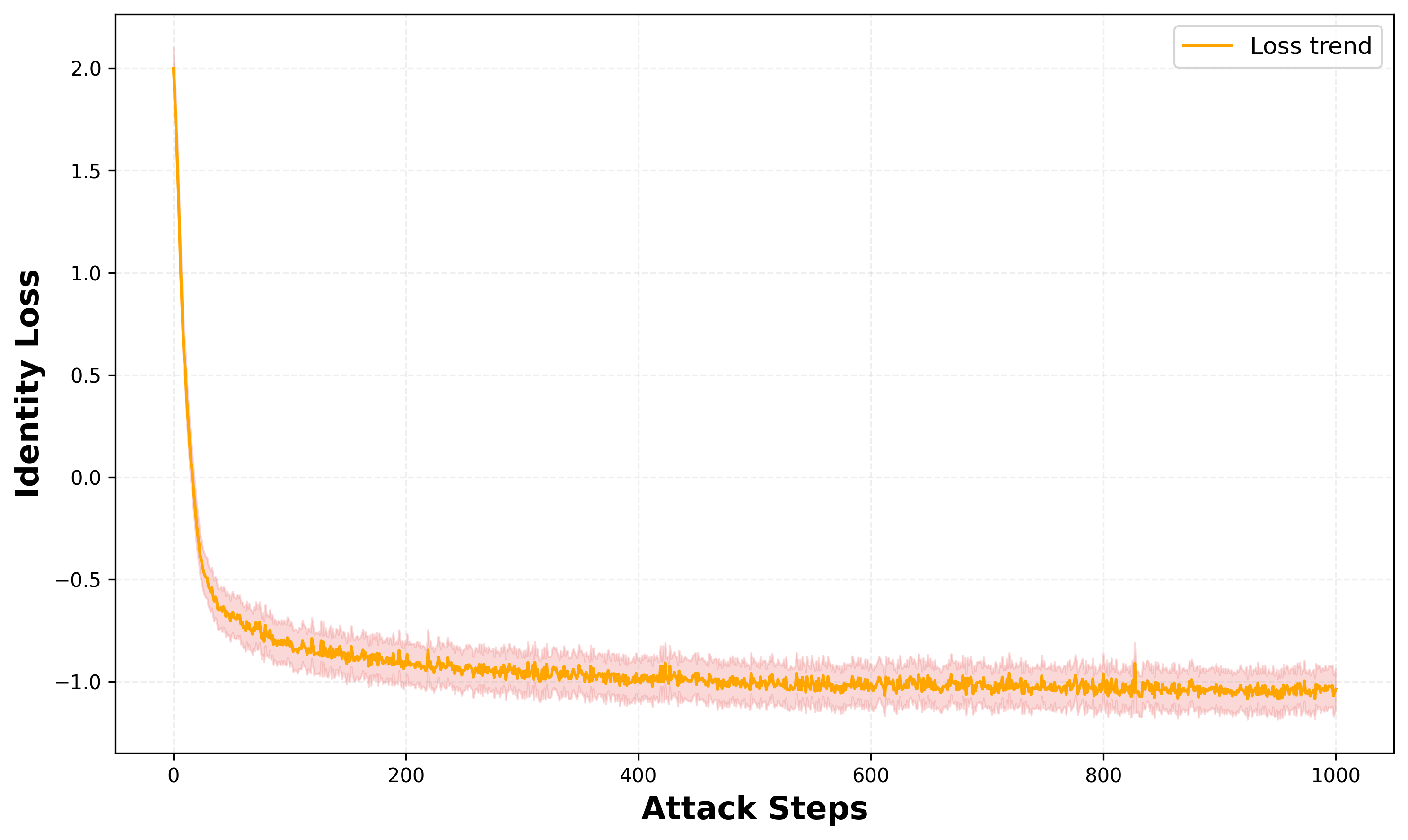}
    \caption{Identity loss convergence over optimization steps with shaded band indicating fluctuation bounds (\(\pm 0.1\)) around the mean curve. The identity loss stabilizes after approximately 200 steps, guiding the selection of the default iteration count.}
    \label{fig:Loss}
\end{figure}

We investigate the convergence behavior of identity loss over a range of optimization steps, up to 1000 iterations. As shown in Figure~\ref{fig:Loss}, the identity loss declines rapidly during the initial 100 steps, indicating early-stage effectiveness in suppressing identity-related features.
To illustrate optimization dynamics more clearly, we apply a shaded band using \texttt{fill\_between}\footnote{\texttt{fill\_between} is a Matplotlib function that fills the area between two curves, allowing visual emphasis on regions such as fluctuation bounds. In this context, it is used to highlight the range of ±0.1 around the loss curve to help visualize the stability and oscillation of the optimization process.} to visualize artificial fluctuation bounds around the loss curve. This is not a confidence interval but a heuristic indicator of stability and oscillation patterns.

The loss saturates around 200 iterations, beyond which further gains become marginal. Based on this trend, we adopt $N_{\text{iter}} = 200$ as the default number of steps in our experiments, ensuring adequate convergence without excessive computation.

\begin{figure}[htbp]
    \centering
    \includegraphics[width=0.47\textwidth]{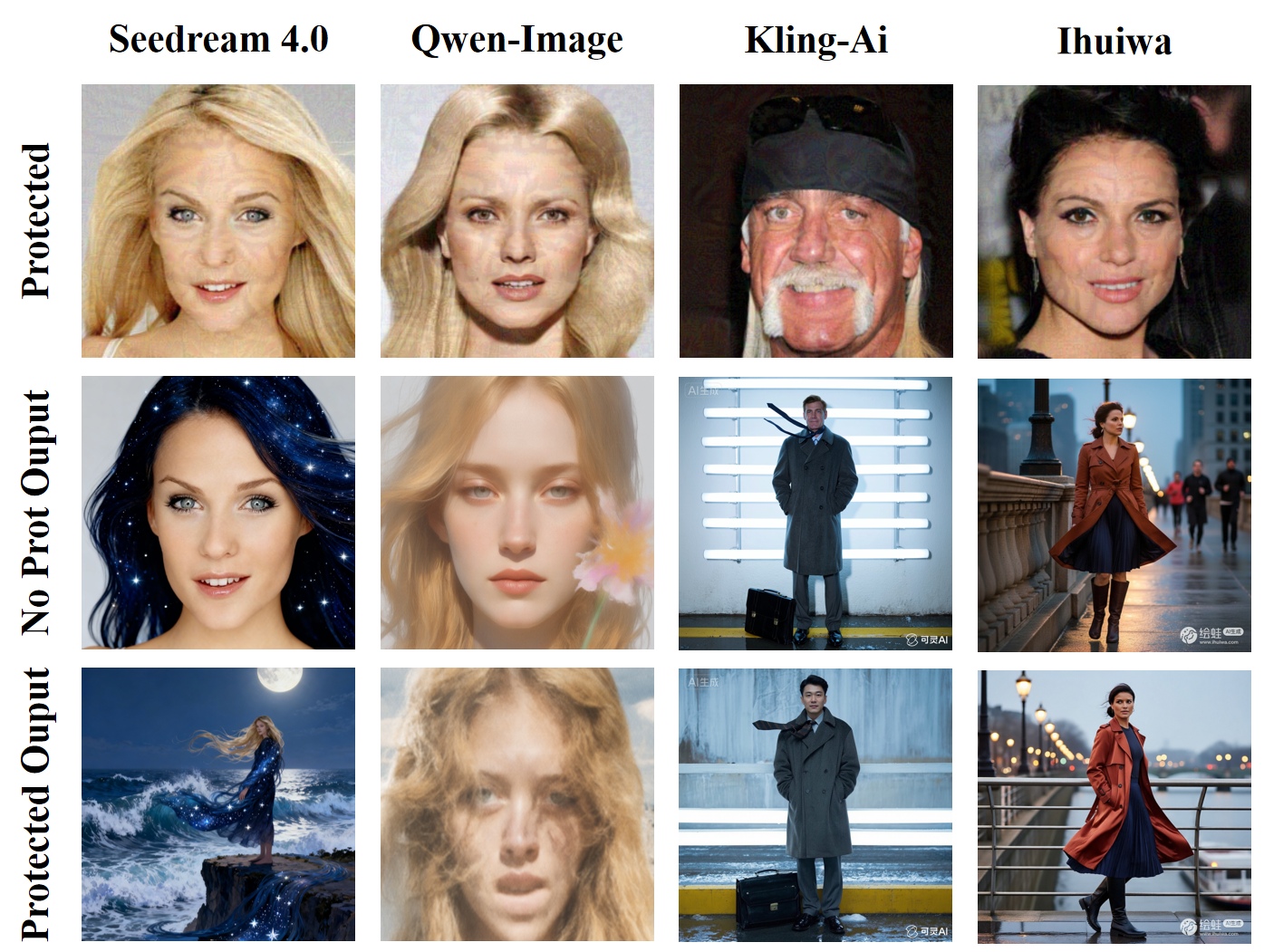}
\caption{Visualization of generalization results on commercial face personalization APIs.}
    \label{fig:api_generalization}
\end{figure}

\section{Generalization Analysis}

\subsection{Evaluation on Commercial Face Personalization APIs}
\label{app:D.1}

To further validate the generalization ability of our protection method, we conducted additional evaluations on several commercial face personalization APIs, including \textit{\textbf{Seedream 4.0}}, \textit{\textbf{Owen-Image}}, \textit{\textbf{Kling-AI}}, and \textit{\textbf{Ihuiwa}}. 
As shown in Figure~\ref{fig:api_generalization}, our method still provides a certain degree of defense under these black-box conditions, demonstrating its robustness and transferability beyond the training environment.

\subsection{Flexibility in Surrogate Selection}
\label{app:D.2}

To evaluate the generalizability of our approach at the model level, we replace the surrogate model used for adversarial optimization with a significantly different generative backbone.  
Our main experiments utilize the IP-Adapter SDXL Plus Face model based on Stable Diffusion XL (SDXL). Additionally, we assess our method on the earlier Stable Diffusion 1.5 (SD1.5) backbone along with its corresponding IP-Adapter variant.  
All attack configurations, including text prompts and optimization parameters, are held constant to ensure comparability.  

\begin{figure*}[htbp]
    \centering
    \includegraphics[width=1\textwidth]{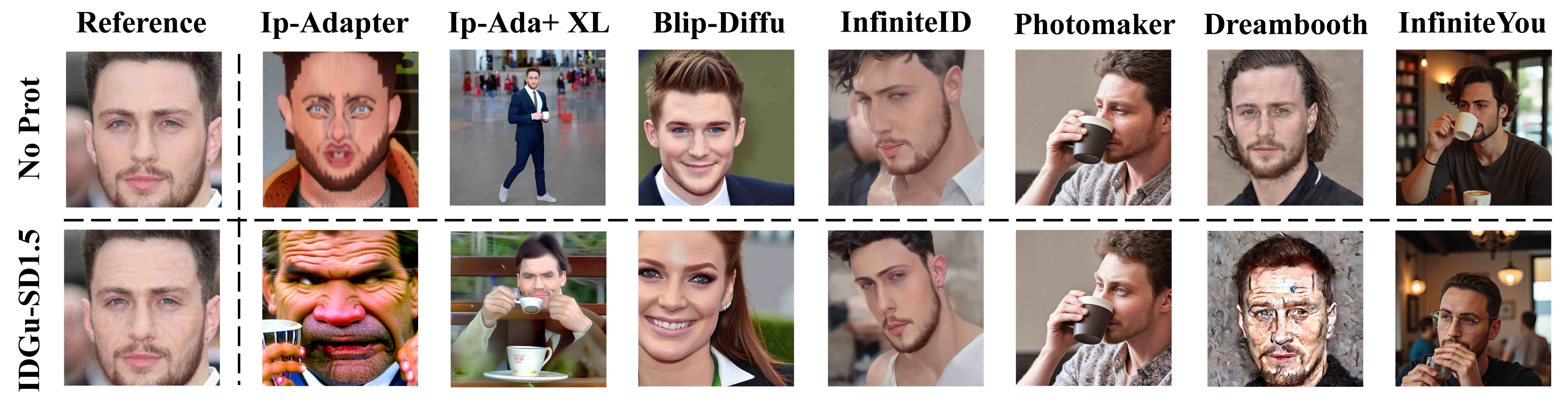}
    \caption{Visual comparison of identity suppression results across different surrogate generative models.}
    \label{fig:vis1}
\end{figure*}

The results demonstrate that the proposed method consistently achieves strong identity suppression across distinct surrogate models, confirming its robustness and applicability beyond specific model architectures, as summarized in Table~\ref{tab:surrogate_model_identity}. Representative visual results further illustrate the consistency and effectiveness of identity suppression across models, as shown in Figure~\ref{fig:vis1}.

\subsection{Dataset Generalization}
\label{app:D.3}

In addition to the VGGFace2 dataset results presented in the main paper, we further evaluate the method on the high-resolution CelebA-HQ dataset.  
We employ multiple identity similarity metrics, namely ArcFace, FaceNet, and VGGFace, to measure the extent of identity removal across diverse personalized generation models.  
Perceptual quality is concurrently assessed through FaceQnet and MagFace to verify preservation of visual fidelity.

The quantitative results summarized in Table~\ref{tab:512 performance_comparison} and Table~\ref{tab:512 Performance.} indicate that the method maintains effective identity suppression and reduced perceptual quality on CelebA-HQ, thus demonstrating robustness across datasets varying in resolution and demographic composition. As illustrated in Figure~\ref{fig:vis3}, the visual results confirm that the method successfully removes identity-specific features while preserving photorealism, supporting the effectiveness of the proposed defense.
We further provide results on the LFW dataset using the identity similarity metric in Table~\ref{tab:R3_lfw_arc_only}. The results remain consistent with those on VGGFace2 and CelebA-HQ, further demonstrating the robustness and cross-dataset generalization ability of IDGuardian.
Collectively, these experiments verify that IDGuardian consistently preserves identity suppression capability and visual quality across different surrogate model architectures and datasets, highlighting its strong generalization capability in diffusion-based personalized face generation.

\begin{figure*}[htbp]
    \centering
    \includegraphics[width=1\textwidth]{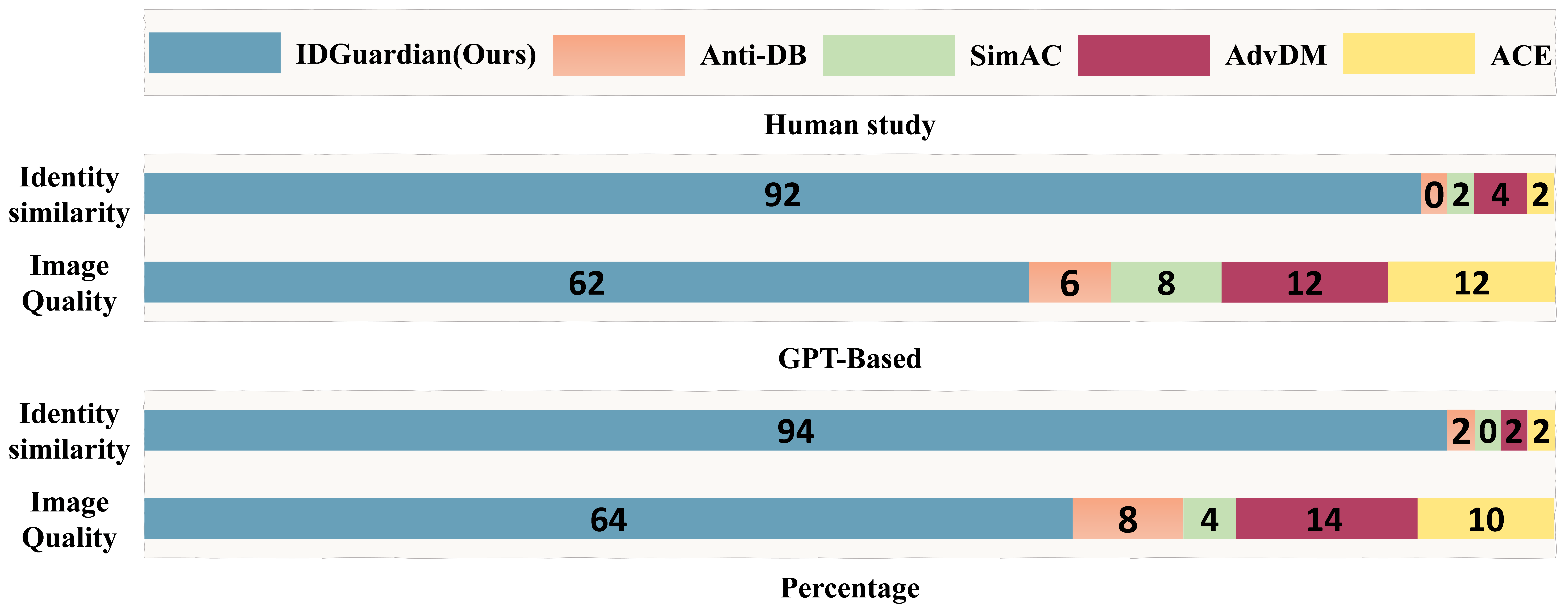}
    \caption{Comparison of Identity Protection Effectiveness Based on Human Perception and GPT-Based Evaluation.}
    \label{fig:human_gpt_eval}
\end{figure*}

\section{Human and GPT-Based Evaluation of Identity Protection Effectiveness}

As illustrated in Figure~\ref{fig:human_gpt_eval}, the upper part of the figure presents the results of the human perceptual study, while the lower part corresponds to the GPT-based evaluation. This joint visualization highlights the consistency between subjective human judgments and automated assessments, with our method favored in 92\% of human responses and 94\% of GPT comparisons across over 50 test cases, further reinforcing the effectiveness of our approach.

\subsection{Human Study}
\label{app:E.1}

To further assess the effectiveness of identity protection from a human perceptual perspective, we conducted a comprehensive subjective evaluation involving human participants. These participants were tasked with rating the perceived identity similarity between the original identity images and the personalized images generated by our method as well as several state-of-the-art baseline methods. The results consistently demonstrate that images produced by our approach exhibit significantly lower perceived identity similarity compared to those generated by other methods, thereby indicating superior identity obfuscation. This extensive subjective evaluation complements the quantitative metrics by capturing nuanced perceptual differences and providing insights into human visual cognition that automated measures alone may not fully capture.

\subsection{GPT-Based Evaluation}
\label{app:E.2}

In addition to the human evaluation, we also employed a state-of-the-art large multimodal language model, GPT, to automatically assess the identity protection performance. Given pairs of original and generated images, GPT was prompted to analyze and identify which generated image exhibits the greatest discrepancy in identity features relative to the original input. Leveraging GPT’s advanced visual-language reasoning capabilities, this automated evaluation approach offers a scalable and efficient complement to human judgment. The findings from the GPT-based evaluation are consistent with the results obtained from the user study, thereby further validating the robustness and effectiveness of our method in substantially reducing identity similarity across different evaluation paradigms.

\subsection{Ethics Statement}
\label{app:E.3}

Our user study involved voluntary participation of adult subjects recruited within our institution. Participants were asked to provide subjective ratings of visual similarity without disclosing or providing any personally identifiable or sensitive information. The study used publicly available face datasets and collected no new identifiable personal data. All user evaluations were performed on synthetic or anonymized data, and no sensitive or personally identifiable information was recorded. The study posed no potential harm or risk to participants. While formal Institutional Review Board (IRB) approval was not required under our institution’s policy for this low-risk, anonymized perceptual evaluation, all participants were informed about the study’s purpose and provided consent prior to participation. The study was conducted in accordance with recognized ethical research principles.

\begin{table*}[htbp]
\centering
\caption{Comparison of identity protection performance against baseline methods.
Evaluation is conducted using three different identity similarity metrics: ArcFace, FaceNet, and VGGFace. 
“No Protected” denotes generation without defense.}
\renewcommand{\arraystretch}{1.5}
\setlength{\tabcolsep}{1.26mm}
\begin{tabular}{lccccccccccccccccccccccc}
\specialrule{1.2pt}{0pt}{0pt} 
\multirow{2}{*}{\textbf{Method}} & \multicolumn{3}{c}{\textbf{\textbf{\textbf{Ip-Adapter}}
}} & \multicolumn{3}{c}{\textbf{\textbf{\textbf{Ip-Adapter Plus XL}}}} & \multicolumn{3}{c}{\textbf{\textbf{\textbf{Blip-Diffusion}}}} & \multicolumn{3}{c}{\textbf{\textbf{\textbf{InfiniteID}}}} \\
\cmidrule(lr){2-4} \cmidrule(lr){5-7} \cmidrule(lr){8-10} \cmidrule(lr){11-13}
&\textbf{Arc $\downarrow$} & \textbf{Face $\downarrow$} & \textbf{VGG $\downarrow$}&\textbf{Arc $\downarrow$} & \textbf{Face $\downarrow$} & \textbf{VGG $\downarrow$} &\textbf{Arc $\downarrow$} & \textbf{Face $\downarrow$} & \textbf{VGG $\downarrow$}&\textbf{Arc $\downarrow$} & \textbf{Face $\downarrow$} & \textbf{VGG $\downarrow$} \\
\specialrule{1.0pt}{0pt}{0pt}
\textbf{No Protected}  & 0.3369  & 0.4774 & 0.3308 &0.4985 & 0.6899 & 0.5573 &0.3305 & 0.3918 & 0.2848 & 0.6345 & 0.7627 & 0.6286 \\
\textbf{Anti-DB}  & 0.2525 & 0.3964 & 0.2715 & 0.4815 & 0.6419  & 0.5219 & 0.2685 & 0.3153 & 0.2286 & 0.5897 & 0.6859 & 0.5848  \\
\textbf{SimAC} & 0.2048 & 0.4005 &	0.2792 & 0.3454 & 0.4634 & 0.3956 & 0.3238 & 0.3344 & 0.2563 & 0.5472 & 0.6670 & 0.5599 \\
\textbf{AdvDM} & 0.2409 & 0.3624 & 0.2535 & 0.3427 & 0.4810 & 0.3700 & 0.2461 & 0.2989 & 0.2363 & 0.5069 & 0.6107 & 0.5087  \\
\textbf{ACE} & 0.2573 & 0.4249 &	0.2996 & 0.4615 & 0.6377 & 0.5079 & 0.2738 & 0.3160 & 0.2336 & 0.5419 & 0.6489 & 0.5383 \\
\textbf{IDProtector} & 0.2059 & 0.4167 &	0.2657 & 0.3324 & 0.5297 & 0.4120 & 0.2565 & 0.2967 & 0.2346 & 0.4972 & 0.7009 & 0.6112 \\
\textbf{IDGuar (Ours)} & \textbf{0.0673} & \textbf{-0.0226} & \textbf{0.0640 }& \textbf{0.1296} & \textbf{0.1231} & \textbf{0.1260} & \textbf{0.2255} & \textbf{0.2376 }& \textbf{0.1964} & \textbf{0.4121} & \textbf{0.5215} & \textbf{0.3990} \\
\specialrule{1.2pt}{0pt}{0pt}
\multirow{2}{*}{\textbf{Method}}  & \multicolumn{3}{c}{\textbf{\textbf{\textbf{Photomaker}}}} & \multicolumn{3}{c}{\textbf{\textbf{\textbf{Dreambooth}}}} & \multicolumn{3}{c}{\textbf{\textbf{\textbf{InfiniteYou}}}} & \multicolumn{3}{c}{\textbf{\textbf{\textbf{Mean}}}} \\
\cmidrule(lr){2-4} \cmidrule(lr){5-7} \cmidrule(lr){8-10} \cmidrule(lr){11-13}
&\textbf{Arc $\downarrow$} & \textbf{Face $\downarrow$} & \textbf{VGG $\downarrow$}&\textbf{Arc $\downarrow$} & \textbf{Face $\downarrow$} & \textbf{VGG $\downarrow$} &\textbf{Arc $\downarrow$} & \textbf{Face $\downarrow$} & \textbf{VGG $\downarrow$}&\textbf{Arc $\downarrow$} & \textbf{Face $\downarrow$}  & \textbf{VGG $\downarrow$} \\
\specialrule{1.0pt}{0pt}{0pt}
\textbf{No Protected}  &0.3488 & 0.3513 & 0.2806 & 0.4681 & 0.6580 & 0.4477 & 0.6788 & 0.7554 & 0.5847 & 0.4709 & 0.5838 & 0.4449\\
\textbf{Anti-DB}   & 0.2331 & 0.3104 & 0.2716 & 0.3590 & 0.5420 & 0.3976 & 0.5867 & 0.6569 & 0.5343 & 0.3959 & 0.5070 & 0.4014 \\
\textbf{SimAC}  & 0.2267 & 0.2614 & 0.2564 & 0.2987 & 0.4760 & 0.3942 & 0.5604 & 0.6472 & 0.5039 & 0.3581 & 0.4643 & 0.3780\\
\textbf{AdvDM}  & 0.2083 & 0.2189 & 0.2477 &\textbf{ 0.2191}& 0.4507 & \textbf{0.3327} & 0.5546 & 0.6309 & 0.4905 & 0.3312 & 0.4362 & 0.3485 \\
\textbf{ACE}  & 0.2246 & 0.2545 & 0.2756 & 0.3252 & 0.5328 & 0.4033  & 0.6450 & 0.7207 & 0.5504 & 0.3900 & 0.5051 & 0.4012\\
\textbf{IDProtector} & 0.2170 & 0.2212 &	0.2807 & 0.3372 & 0.4768 & 0.3932 & 0.5047 & 0.6067 & 0.5105 & 0.3358 & 0.4641 & 0.3868 \\
\textbf{IDGuar (Ours)}  & \textbf{0.0747} & \textbf{0.1284} & \textbf{0.2239} & 0.2792 & \textbf{0.3893} & 0.3404 & \textbf{0.3230 }&\textbf{ 0.3265} & \textbf{0.2522} & \textbf{0.2159} & \textbf{0.2434} & \textbf{0.2288}\\
\specialrule{1.2pt}{0pt}{0pt}
\end{tabular}

\label{tab:com performance_comparison}

\end{table*}

\begin{table*}[htbp]
\centering
\caption{Comparison of identity protection performance against baseline methods.
Evaluation is conducted using FaceQNet and MagFace quality scores.
“No Protected” denotes generation without defense.}
\renewcommand{\arraystretch}{1.5}
\setlength{\tabcolsep}{1.06mm}
\begin{tabular}{lccccccccccccccccccc}
\specialrule{1.2pt}{0pt}{0pt} 
    \multirow{2}{*}{\textbf{Method}} & \multicolumn{2}{c}{\textbf{\textbf{\textbf{Ip-Adapter}}}} & \multicolumn{2}{c}{\textbf{\textbf{\textbf{Ip-Adapter Plus XL}}}} & \multicolumn{2}{c}{\textbf{\textbf{\textbf{Blip-Diffusion}}}} & \multicolumn{2}{c}{\textbf{\textbf{\textbf{InfiniteID}}}}   \\
\cmidrule(lr){2-3} \cmidrule(lr){4-5} \cmidrule(lr){6-7} \cmidrule(lr){8-9} 
& \textbf{FaceQNet $\downarrow$} & \textbf{MagFace $\downarrow$} & \textbf{FaceQNet $\downarrow$} & \textbf{MagFace $\downarrow$} &\textbf{FaceQNet $\downarrow$} & \textbf{MagFace $\downarrow$} & \textbf{FaceQNet $\downarrow$} & \textbf{MagFace $\downarrow$} \\
\specialrule{1.0pt}{0pt}{0pt}
\textbf{No Protected}  & 0.3339 & 21.3847 & 0.4564 & 19.3085 & 0.3305 & 21.8112 & 0.4946 & 19.0747   \\
\textbf{Anti-DB}  & 0.3045 & 21.1209 & 0.4553 & \textbf{18.4376} & 0.3565 & 20.8991 & 0.4957 & 18.9697  \\
\textbf{SimAC }& 0.3124 & 20.8217 & 0.4492 & 18.9702 & 0.3628 & 21.4539 & \textbf{0.4844} & 19.0755\\
\textbf{AdvDM} & 0.3486 & 20.6235& \textbf{0.4379} & 18.8501 & 0.3547 & 21.6632 & 0.4896 & 18.9385 \\
\textbf{ACE} & 0.3627 & 20.6235 & 0.4528 & 18.5293 & 0.3630 & 21.7078 & 0.4994 & 19.0961 \\
\textbf{IDProtector} & 0.3367 & \textbf{19.9330} & 0.4318 & 18.8869 & 0.3680 & 21.3936 & 0.4885 & 18.9833 \\
\textbf{IDGuar (Ours)} & \textbf{0.2820} & 20.8712 & 0.4503 & 18.6099 & \textbf{0.3511} & \textbf{20.8498} & 0.4938 & \textbf{18.9031}  \\
\specialrule{1.2pt}{0pt}{0pt}
    \multirow{2}{*}{\textbf{Method}}  & \multicolumn{2}{c}{\textbf{\textbf{\textbf{Photomaker}}}} & \multicolumn{2}{c}{\textbf{\textbf{\textbf{Dreambooth}}}} & \multicolumn{2}{c}{\textbf{\textbf{\textbf{InfiniteYou}}}} & \multicolumn{2}{c}{\textbf{\textbf{\textbf{Mean}}}}  \\
\cmidrule(lr){2-3} \cmidrule(lr){4-5} \cmidrule(lr){6-7} \cmidrule(lr){8-9} 
& \textbf{FaceQNet $\downarrow$} & \textbf{MagFace $\downarrow$} & \textbf{FaceQNet $\downarrow$} & \textbf{MagFace $\downarrow$} &\textbf{FaceQNet $\downarrow$} & \textbf{MagFace $\downarrow$} & \textbf{FaceQNet $\downarrow$} & \textbf{MagFace $\downarrow$}  \\
\specialrule{1.0pt}{0pt}{0pt}
\textbf{No Protected}   & 0.4564 & 20.0541 & 0.3316 & 20.2766 & 0.4927 & 20.7298 & 0.4137 & 20.3771\\
\textbf{Anti-DB}  & 0.2883 & 20.0427 & 0.3186 & 20.2498 & 0.4666 & 19.9501 & 0.3836 & 19.9529 \\
\textbf{SimAC } & 0.2959 & 19.9726 & 0.3427 & 20.1656 & 0.4590 & 19.9077 & 0.3866 & 20.0525 \\
\textbf{AdvDM}  & 0.2934 & 20.1307 & \textbf{0.1921} & \textbf{19.7116} & 0.4586 & 19.9321 & 0.3678 & 19.9785 \\
\textbf{ACE}  & 0.2866 & 20.1201 & 0.2147 & 20.1827 & 0.4610 & 19.8904 & 0.3772 & 19.8786 \\
\textbf{IDProtector} & 0.3119 & 28.5475 & 0.2248 & 20.6773 & 0.4535 & 19.5957 & 0.3736 & 21.1453 \\
\textbf{IDGuar (Ours)} & \textbf{0.2733} & \textbf{19.8202} & 0.2203 & 20.2199 & \textbf{0.4517} & \textbf{19.8111}  & \textbf{0.3604} & \textbf{19.8693}\\
\specialrule{1.2pt}{0pt}{0pt}
\end{tabular}

\label{tab:com performance_comparison1}

\end{table*}

\begin{table*}[htbp]
\centering
\caption{Ablation study of identity loss combinations and guidance directions, where \textbf{G adv} denotes guidance solely toward the adversarial identity distribution and \textbf{G clean} denotes guidance solely away from the clean identity distribution.}
\renewcommand{\arraystretch}{1.5}
\setlength{\tabcolsep}{1.26mm}
\begin{tabular}{lcccccccccccccccccccccc}
\specialrule{1.2pt}{0pt}{0pt} 
\multirow{2}{*}{\textbf{Method}} & \multicolumn{3}{c}{\textbf{\textbf{\textbf{Ip-Adapter}}
}} & \multicolumn{3}{c}{\textbf{\textbf{\textbf{Ip-Adapter Plus XL}}}} & \multicolumn{3}{c}{\textbf{\textbf{\textbf{Blip-Diffusion}}}} & \multicolumn{3}{c}{\textbf{\textbf{\textbf{InfiniteID}}}} \\
\cmidrule(lr){2-4} \cmidrule(lr){5-7} \cmidrule(lr){8-10} \cmidrule(lr){11-13} 
&\textbf{Arc $\downarrow$} & \textbf{Face $\downarrow$} & \textbf{VGG $\downarrow$}&\textbf{Arc $\downarrow$} & \textbf{Face $\downarrow$} & \textbf{VGG $\downarrow$} &\textbf{Arc $\downarrow$} & \textbf{Face $\downarrow$} & \textbf{VGG $\downarrow$}&\textbf{Arc $\downarrow$} & \textbf{Face $\downarrow$} & \textbf{VGG $\downarrow$} \\
\specialrule{1.0pt}{0pt}{0pt}
\textbf{No Protected}  & 0.3369  & 0.4774 & 0.3308 &0.4985 & 0.6899 & 0.5573 &0.3305 & 0.3918 & 0.2848 & 0.6345 & 0.7627 & 0.6286 \\
\specialrule{0.8pt}{0pt}{0pt}
\textbf{CLIP}  & 0.2289 & 0.3058 & 0.1949 & 0.2032 & 0.2328 & 0.2147 & 0.2864 & 0.3078 & 0.2664  & 0.5048	 & 0.6775 & 0.5478\\
\textbf{Facenet}  & 0.2412 & 0.2897 & 0.2058 & 0.2145 & 0.2281 & 0.2223 & 0.2948 & 0.2989 & 0.2716 & 0.5126 & 0.6843 & 0.5534\\
\textbf{CLIP + Facenet} & 0.1578 & 0.2347 & 0.1326 & 0.1665 & 0.1707 & 0.1638 & 0.2534 & 0.2764 & 0.2355 & 0.4932 & 0.6047 & 0.5089  \\
\textbf{IDGuar (Ours)} & \textbf{0.0673} & \textbf{-0.0226} & \textbf{0.0640 }& \textbf{0.1296} & \textbf{0.1231} & \textbf{0.1260} & \textbf{0.2255} & \textbf{0.2376 }& \textbf{0.1964} & \textbf{0.4121} & \textbf{0.5215} & \textbf{0.3990}  \\
\specialrule{1.0pt}{0pt}{0pt}
\textbf{G adv} & 0.0688 & -0.0006 & 0.0650 & 0.1318 & 0.1476 & 0.1536 & 0.2971 & 0.3072 & 0.2509 & 0.4501 & 0.5369 & 0.4181  \\
\textbf{G clean} & 0.0869 & -0.0061 & 0.0668 & 0.1442 & 0.1274 & 0.1432 & 0.2400 & 0.2751 & 0.2029 & 0.4265 & 0.5261 & 0.4073   \\
\textbf{IDGuar (Ours)} & \textbf{0.0673} & \textbf{-0.0226} & \textbf{0.0640 }& \textbf{0.1296} & \textbf{0.1231} & \textbf{0.1260} & \textbf{0.2255} & \textbf{0.2376 }& \textbf{0.1964} & \textbf{0.4121} & \textbf{0.5215} & \textbf{0.3990}  \\
\specialrule{1.2pt}{0pt}{0pt} 
\multirow{2}{*}{\textbf{Method}} & \multicolumn{3}{c}{\textbf{\textbf{\textbf{Photomaker}}}} & \multicolumn{3}{c}{\textbf{\textbf{\textbf{Dreambooth}}}} & \multicolumn{3}{c}{\textbf{\textbf{\textbf{InfiniteYou}}}} & \multicolumn{3}{c}{\textbf{\textbf{\textbf{Mean}}}} \\
\cmidrule(lr){2-4} \cmidrule(lr){5-7} \cmidrule(lr){8-10} \cmidrule(lr){11-13}
&\textbf{Arc $\downarrow$} & \textbf{Face $\downarrow$} & \textbf{VGG $\downarrow$}&\textbf{Arc $\downarrow$} & \textbf{Face $\downarrow$} & \textbf{VGG $\downarrow$} &\textbf{Arc $\downarrow$} & \textbf{Face $\downarrow$} & \textbf{VGG $\downarrow$}&\textbf{Arc $\downarrow$} & \textbf{Face $\downarrow$} & \textbf{VGG $\downarrow$} \\
\specialrule{1.0pt}{0pt}{0pt} 
\textbf{No Protected} &0.3488 & 0.3513 & 0.2806 & 0.4681 & 0.6580 & 0.4477  & 0.6788 & 0.7554 & 0.5847 & 0.4709 & 0.5838 & 0.4449\\
\specialrule{0.8pt}{0pt}{0pt}
\textbf{CLIP}  & 0.2465 & 0.2696 & 0.2757 & 0.4634 & 0.5890 & 0.4584 & 0.5273 & 0.6043 & 0.4599 & 0.3458 & 0.4210 & 0.3397 \\
\textbf{Facenet}  & 0.2593 & 0.2634 & 0.2810 & 0.4721 & 0.5743 & 0.4628 & 0.5345 & 0.5926 & 0.4658 & 0.3612 & 0.4188 & 0.3518\\
\textbf{CLIP + Facenet}  & 0.2056 & 0.2142 & 0.2202 & 0.4577 & 0.5606 & 0.4726 & 0.5032 & 0.5357 & 0.3932 & 0.3063 & 0.3577 & 0.2905\\
\textbf{IDGuar (Ours)}& \textbf{0.0747} & \textbf{0.1284} & \textbf{0.2239} & \textbf{0.2792} & \textbf{0.3893} & \textbf{0.3404} & \textbf{0.3230} & \textbf{0.3265 }& \textbf{0.2522}  & \textbf{0.2159} & \textbf{0.2434} & \textbf{0.2288}\\
\specialrule{1.0pt}{0pt}{0pt}
\textbf{G adv}  & 0.1437 & 0.1477 & 0.2324  & 0.2796  & 0.4187  & 0.3432  & 0.3279  & 0.3405  & 0.2526 & 0.2427 & 0.2711 & 0.2451\\
\textbf{G clean}  & 0.1562 & 0.1974 & 0.2333  & 0.3046 & 0.4813  & 0.3410  & 0.3379  & 0.3472  & 0.2565 & 0.2423 & 0.2783 & 0.2358 \\
\textbf{IDGuar (Ours)}& \textbf{0.0747} & \textbf{0.1284} & \textbf{0.2239} & \textbf{0.2792} & \textbf{0.3893} & \textbf{0.3404} & \textbf{0.3230} & \textbf{0.3265 }& \textbf{0.2522}  & \textbf{0.2159} & \textbf{0.2434} & \textbf{0.2288}\\
\specialrule{1.2pt}{0pt}{0pt} 
\end{tabular}

\label{tab:Ablation performance_comparison}
\end{table*}

\begin{table*}[htbp]
\centering
\caption{Ablation study of different identity loss combinations and guidance directions. Evaluation is conducted using FaceQNet and MagFace to measure the perceptual quality of generated images.}
\renewcommand{\arraystretch}{1.5}
\setlength{\tabcolsep}{1.0mm}
\begin{tabular}{lccccccccccccccccccc}
\specialrule{1.2pt}{0pt}{0pt} 
\multirow{2}{*}{\textbf{Method}} & \multicolumn{2}{c}{\textbf{\textbf{\textbf{Ip-Adapter}}}} & \multicolumn{2}{c}{\textbf{\textbf{\textbf{Ip-Adapter Plus XL}}}} & \multicolumn{2}{c}{\textbf{\textbf{\textbf{Blip-Diffusion}}}} & \multicolumn{2}{c}{\textbf{\textbf{\textbf{InfiniteID}}}} \\
\cmidrule(lr){2-3} \cmidrule(lr){4-5} \cmidrule(lr){6-7} \cmidrule(lr){8-9} 
& \textbf{FaceQNet $\downarrow$} & \textbf{MagFace $\downarrow$} & \textbf{FaceQNet $\downarrow$} & \textbf{MagFace $\downarrow$} &\textbf{FaceQNet $\downarrow$} & \textbf{MagFace $\downarrow$} & \textbf{FaceQNet $\downarrow$} & \textbf{MagFace $\downarrow$} \\
\specialrule{1.0pt}{0pt}{0pt}
\textbf{No Protected}  & 0.3339 & 21.3847 & 0.4564 & 19.3085 & 0.3305 & 21.8112 & 0.4946 & 19.0747 \\
\specialrule{0.8pt}{0pt}{0pt}
\textbf{CLIP} & 0.3129 & 21.7832 & 0.4726 & 20.4175 & 0.3683 & 20.9487 & 0.5142 & 19.6254  \\
\textbf{Facenet} & 0.3185 & 21.6910 & 0.4791  & 20.5083 & 0.3724 & \textbf{20.0284}  & 0.5076 & 19.4939  \\
\textbf{CLIP + Facenet} & \textbf{0.2714} & 21.8426 & 0.4689 & 19.5268 & 0.3776 & 20.4123 & 0.5128 & 20.1884
\\
 \textbf{IDGuar (Ours)} & 0.2820 & \textbf{20.8712} & \textbf{0.4503} & \textbf{18.6099} & \textbf{0.3511} & 20.8498 & \textbf{0.4938} & \textbf{18.8031}  \\
 \specialrule{1.0pt}{0pt}{0pt}
\textbf{G adv} & 0.2900 & \textbf{20.8653} & 0.4603 & 19.2300 & 0.3569 & 20.7490 & \textbf{0.4899} & 19.0814 \\
\textbf{G clean} &\textbf{0.2692}& 21.6448 & 0.4526 & 19.2488 & 0.3557 & \textbf{19.3413 }& 0.4968 & 18.8645  \\
\textbf{IDGuar (Ours)} & 0.2820 & 20.8712 & \textbf{0.4503} & \textbf{18.6099} & \textbf{0.3511} & 20.8498 & 0.4938 & \textbf{18.8031}  \\
\specialrule{1.2pt}{0pt}{0pt}
\multirow{2}{*}{\textbf{Method}} & \multicolumn{2}{c}{\textbf{\textbf{\textbf{Photomaker}}}} & \multicolumn{2}{c}{\textbf{\textbf{\textbf{Dreambooth}}}} & \multicolumn{2}{c}{\textbf{\textbf{\textbf{InfiniteYou}}}} & \multicolumn{2}{c}{\textbf{\textbf{\textbf{Mean}}}} \\
\cmidrule(lr){2-3} \cmidrule(lr){4-5} \cmidrule(lr){6-7} \cmidrule(lr){8-9} 
& \textbf{FaceQNet $\downarrow$} & \textbf{MagFace $\downarrow$} & \textbf{FaceQNet $\downarrow$} & \textbf{MagFace $\downarrow$} &\textbf{FaceQNet $\downarrow$} & \textbf{MagFace $\downarrow$} & \textbf{FaceQNet $\downarrow$} & \textbf{MagFace $\downarrow$} \\
\specialrule{1.0pt}{0pt}{0pt}
\textbf{No Protected}  & 0.4564 & 20.0541 & 0.3316 & 20.2766 & 0.4527 & 19.7298 & 0.4083 & 20.2342 \\
\specialrule{0.8pt}{0pt}{0pt}
\textbf{CLIP} & 0.3012 & 20.2741 & 0.3310 & \textbf{19.5135} & 0.4729 & \textbf{19.1276} & 0.3962 & 20.2414\\
\textbf{Facenet} & 0.3075 & 20.3417 & 0.3298 & 19.8824 & 0.4660  & 19.4059 & 0.3972 & 20.1930  \\
\textbf{CLIP + Facenet}  & 0.2916 & 20.1042 & 0.2983 & 19.8631 & 0.4679 & 20.0893 & 0.3768 & 20.2895 \\
\textbf{IDGuar (Ours)}  & \textbf{0.2733} & \textbf{19.8202}& \textbf{0.2203} & 20.2199 & \textbf{0.4517 }& 19.8111 & \textbf{0.3604 }& \textbf{19.8693}\\
\specialrule{1.0pt}{0pt}{0pt}
\textbf{G adv}  & 0.2813 & \textbf{19.7368} & 0.2754 & \textbf{19.9528} & 0.4536 & 20.2709 & 0.3725 & 19.9837\\
\textbf{G clean}  & 0.2789 & 19.8856 & 0.2270 & 20.5887 & 0.4519 & 20.3764 & 0.3617 & 19.9929 \\
\textbf{IDGuar (Ours)}  & \textbf{0.2733} & 19.8202 & \textbf{0.2203} & 20.2199 & \textbf{0.4517 }& \textbf{19.8111} & \textbf{0.3604 }& \textbf{19.8693}\\
\specialrule{1.2pt}{0pt}{0pt}
\end{tabular}

\label{tab:Ablation performance_comparison1}
\end{table*}

\begin{table*}[htbp]
\centering
\caption{Robustness evaluation against common image post-processing operations, including JPEG compression, Gaussian blur, and Gaussian noise, and attacks targeting active defenses such as Impress.}
\renewcommand{\arraystretch}{1.4}
\setlength{\tabcolsep}{1.26mm}
\begin{tabular}{lcccccccccccccccccccccc}
\specialrule{1.2pt}{0pt}{0pt} 
    \multirow{2}{*}{\textbf{Method}} & \multicolumn{3}{c}{\textbf{\textbf{\textbf{Ip-Adapter}}
}} & \multicolumn{3}{c}{\textbf{\textbf{\textbf{Ip-Adapter Plus XL}}}} & \multicolumn{3}{c}{\textbf{\textbf{\textbf{Blip-Diffusion}}}} & \multicolumn{3}{c}{\textbf{\textbf{\textbf{InfiniteID}}}}  \\
\cmidrule(lr){2-4} \cmidrule(lr){5-7} \cmidrule(lr){8-10} \cmidrule(lr){11-13} 
&\textbf{Arc $\downarrow$} & \textbf{Face $\downarrow$} & \textbf{VGG $\downarrow$}&\textbf{Arc $\downarrow$} & \textbf{Face $\downarrow$} & \textbf{VGG $\downarrow$} &\textbf{Arc $\downarrow$} & \textbf{Face $\downarrow$} & \textbf{VGG $\downarrow$}&\textbf{Arc $\downarrow$} & \textbf{Face $\downarrow$} & \textbf{VGG $\downarrow$} \\
\specialrule{1.0pt}{0pt}{0pt}
\textbf{No Protected}  & 0.3369  & 0.4774 & 0.3308 &0.4985 & 0.6899 & 0.5573 &0.3305 & 0.3918 & 0.2848 & 0.6345 & 0.7627 & 0.6286 \\
\textbf{JPEG}  & 0.1068 & 0.1900 & 0.1263 & 0.3209 & 0.3983 & 0.3006 & 0.2236 & 0.2196 & 0.2026 & 0.4286 & 0.5158 & 0.4062   \\
\textbf{Blur} & 0.1678 & 0.2948 & 0.1967 & 0.3317 & 0.4443 & 0.3520 & 0.2663 & 0.2663 & 0.2077 & \textbf{0.3854} & \textbf{0.4780} & \textbf{0.3734}  \\
\textbf{Noise} & 0.1365 & 0.1794 & 0.1295 & 0.3078 & 0.3970 & 0.3113 & \textbf{0.1916 }& \textbf{0.1910} &\textbf{0.1716} & 0.4482 & 0.5581 & 0.4391 \\
\textbf{Impress} & 0.1018 & 0.1696 & 0.1230 & 0.3184 & 0.3735 & 0.2956 & 0.2261 & 0.2653 & 0.2277 & 0.4478 & 0.5309 & 0.4074 \\
\textbf{IDGuar (Ours)} & \textbf{0.0673} & \textbf{-0.0226} & \textbf{0.0640 }& \textbf{0.1296} & \textbf{0.1231} & \textbf{0.1260} & 0.2255 & 0.2376 & 0.1964 & 0.4121 & 0.5215 & 0.3990\\
\specialrule{1.2pt}{0pt}{0pt}
    \multirow{2}{*}{\textbf{Method}}  & \multicolumn{3}{c}{\textbf{\textbf{\textbf{Photomaker}}}} & \multicolumn{3}{c}{\textbf{\textbf{\textbf{Dreambooth}}}} & \multicolumn{3}{c}{\textbf{\textbf{\textbf{InfiniteYou}}}} & \multicolumn{3}{c}{\textbf{\textbf{\textbf{Mean}}}} \\
\cmidrule(lr){2-4} \cmidrule(lr){5-7} \cmidrule(lr){8-10} \cmidrule(lr){11-13} 
&\textbf{Arc $\downarrow$} & \textbf{Face $\downarrow$} & \textbf{VGG $\downarrow$}&\textbf{Arc $\downarrow$} & \textbf{Face $\downarrow$} & \textbf{VGG $\downarrow$} &\textbf{Arc $\downarrow$} & \textbf{Face $\downarrow$} & \textbf{VGG $\downarrow$}&\textbf{Arc $\downarrow$} & \textbf{Face $\downarrow$} & \textbf{VGG $\downarrow$} \\
\specialrule{1.0pt}{0pt}{0pt}
\textbf{No Protected}  &0.3488 & 0.3513 & 0.2806 & 0.4681 & 0.6580 & 0.4477  & 0.6788 & 0.7554 & 0.5847 & 0.4709 & 0.5838 & 0.4449\\
\textbf{JPEG}   & 0.1199 & \textbf{0.0754} & 0.2082 & 0.2723 & 0.3963 & \textbf{0.3267 }& 0.3344 & 0.3728 & 0.2590 & 0.2581 & 0.3097 & 0.2614  \\
\textbf{Blur}  & \textbf{0.0709} & 0.1591 & 0.2102  & 0.2899  & 0.5136 & 0.4110 & 0.3371 & 0.3510 & 0.2587 & 0.2642 & 0.3582 & 0.2871\\
\textbf{Noise}  & 0.1785 & 0.1618 & \textbf{0.1971} & 0.3418 & 0.5177 & 0.3844 & 0.3410 & 0.3625 & 0.2623 & 0.2780 & 0.3382 & 0.2708\\
\textbf{Impress}  & 0.1910 & 0.1494 & 0.2097 &\textbf{ 0.2516} &	0.4479 & 0.3386 & \textbf{0.3166} & \textbf{0.3233} & \textbf{0.2387} & 0.2648 & 0.3229 & 0.2629\\
\textbf{IDGuar (Ours)}  & 0.0747 & 0.1284 & 0.2239 & 0.2792 & \textbf{0.3893} & 0.3404 & 0.3230 & 0.3265 & 0.2522 & \textbf{0.2159} & \textbf{0.2434} & \textbf{0.2288} \\
\specialrule{1.2pt}{0pt}{0pt}
\end{tabular}

\label{tab:Robustness performance_comparison}
\end{table*}

\begin{table*}[htbp]
\centering
\caption{Robustness evaluation against common image post-processing operations. Evaluation is conducted using FaceQNet and MagFace to measure the perceptual quality of generated images.}
\renewcommand{\arraystretch}{1.4}
\setlength{\tabcolsep}{1.06mm}
\begin{tabular}{lccccccccccccccccccc}
\specialrule{1.2pt}{0pt}{0pt} 
    \multirow{2}{*}{\textbf{Method}} & \multicolumn{2}{c}{\textbf{\textbf{\textbf{Ip-Adapter}}}} & \multicolumn{2}{c}{\textbf{\textbf{\textbf{Ip-Adapter Plus XL}}}} & \multicolumn{2}{c}{\textbf{\textbf{\textbf{Blip-Diffusion}}}} & \multicolumn{2}{c}{\textbf{\textbf{\textbf{InfiniteID}}}}  \\
\cmidrule(lr){2-3} \cmidrule(lr){4-5} \cmidrule(lr){6-7} \cmidrule(lr){8-9} 
& \textbf{FaceQNet $\downarrow$} & \textbf{MagFace $\downarrow$} & \textbf{FaceQNet $\downarrow$} & \textbf{MagFace $\downarrow$} &\textbf{FaceQNet $\downarrow$} & \textbf{MagFace $\downarrow$} & \textbf{FaceQNet $\downarrow$} & \textbf{MagFace $\downarrow$} \\
\specialrule{1.0pt}{0pt}{0pt}
\textbf{No Protected}  & 0.3339 & 21.3847 & 0.4564 & 19.3085 & 0.3305 & 21.8112 & 0.4946 & 19.0747  \\
\textbf{JPEG} & \textbf{0.2211} & 21.3679 & 0.4539 & 18.5163 & 0.3233 & 21.7881 & 0.4918 & \textbf{18.8227}  \\
\textbf{Blur} & 0.3211 & 21.3901 & 0.4667 & 19.7750 & \textbf{0.2895} & \textbf{19.8607} & 0.5031 & 18.8229 \\
\textbf{Noise} & 0.2499 & \textbf{20.6822 }& 0.4497 & \textbf{18.4623} & 0.3460 & 20.6286 & 0.5016 & 19.0230 \\
\textbf{Impress} & 0.2501 & 20.7604 & \textbf{0.4270} & 19.6015 & 0.3353 & 20.9642 & 0.4935 & 19.1813\\
\textbf{IDGuar (Ours)} & 0.2820 & 20.8712 & 0.4503 & 18.5099 & 0.3511 & 20.8498 & \textbf{0.4908} & 18.9031 \\
\specialrule{1.2pt}{0pt}{0pt}
\multirow{2}{*}{\textbf{Method}} & \multicolumn{2}{c}{\textbf{\textbf{\textbf{Photomaker}}}} & \multicolumn{2}{c}{\textbf{\textbf{\textbf{Dreambooth}}}} & \multicolumn{2}{c}{\textbf{\textbf{\textbf{InfiniteYou}}}} & \multicolumn{2}{c}{\textbf{\textbf{\textbf{Mean}}}}  \\
\cmidrule(lr){2-3} \cmidrule(lr){4-5} \cmidrule(lr){6-7} \cmidrule(lr){8-9} 
& \textbf{FaceQNet $\downarrow$} & \textbf{MagFace $\downarrow$} & \textbf{FaceQNet $\downarrow$} & \textbf{MagFace $\downarrow$} &\textbf{FaceQNet $\downarrow$} & \textbf{MagFace $\downarrow$} & \textbf{FaceQNet $\downarrow$} & \textbf{MagFace $\downarrow$}  \\
\specialrule{1.0pt}{0pt}{0pt}
\textbf{No Protected}  & 0.4564 & 20.0541 & 0.3316 & 20.2766 & 0.4527 & 19.7298  & 0.4080 & 20.2342\\
\textbf{JPEG} & 0.2743 & \textbf{19.6998}& 0.2535 & 20.5918 & 0.4544 & 20.0803 & \textbf{0.3532} & 19.9809 \\
\textbf{Blur} & 0.2754 & 19.8549 & 0.2814 & 19.6256 & \textbf{0.4483} & 20.0016 & 0.3694 & 19.9044\\
\textbf{Noise} & 0.2809 & 19.8303 & 0.3583 & \textbf{19.2192} & 0.4528 & 19.8165 &  0.3770 & \textbf{19.6660}\\
\textbf{Impress} & 0.2762 & 20.0743 & 0.2475 & 21.0550 & 0.4536 & \textbf{19.6948}  & 0.3547 & 20.1902\\
\textbf{IDGuar (Ours)}  & \textbf{0.2733} & 19.8202 & \textbf{0.2203} & 20.2199 & 0.4517 & 19.8111 & 0.3604 & 19.8693 \\
\specialrule{1.2pt}{0pt}{0pt}
\end{tabular}

\label{tab:Robustness performance_comparison1}
\end{table*}

\begin{table*}[htbp]
\centering
\caption{Comparison of identity protection performance against baseline methods on the CelebA-HQ dataset. Evaluation is conducted using three widely adopted identity similarity metrics: ArcFace, FaceNet, and VGGFace.}
\renewcommand{\arraystretch}{1.5}
\setlength{\tabcolsep}{1.36mm}
\begin{tabular}{lccccccccccccccccccc}
\specialrule{1.2pt}{0pt}{0pt} 
 \multirow{2}{*}{\textbf{Method}} & \multicolumn{3}{c}{\textbf{\textbf{\textbf{Ip-Adapter}}
}} & \multicolumn{3}{c}{\textbf{\textbf{\textbf{Ip-Adapter Plus XL}}}} & \multicolumn{3}{c}{\textbf{\textbf{\textbf{Blip-Diffusion}}}} & \multicolumn{3}{c}{\textbf{\textbf{\textbf{InfiniteID}}}}  \\
\cmidrule(lr){2-4} \cmidrule(lr){5-7} \cmidrule(lr){8-10} \cmidrule(lr){11-13}  
&\textbf{Arc $\downarrow$} & \textbf{Face $\downarrow$} & \textbf{VGG $\downarrow$}&\textbf{Arc $\downarrow$} & \textbf{Face $\downarrow$} & \textbf{VGG $\downarrow$} &\textbf{Arc $\downarrow$} & \textbf{Face $\downarrow$} & \textbf{VGG $\downarrow$}&\textbf{Arc $\downarrow$} & \textbf{Face $\downarrow$} & \textbf{VGG $\downarrow$} \\
\specialrule{1.0pt}{0pt}{0pt}
\textbf{No Protected} & 0.4104 & 0.4714 & 0.3764 & 0.5706 & 0.7152 & 0.6062 & 0.2862 & 0.3495 & 0.3061 & 0.6352 & 0.6886 & 0.5642   \\
\textbf{Anti-DB}  & 0.3061 & 0.3647 & 0.3193 & 0.4940 & 0.6321 & 0.5260 & 0.2257 & 0.3369 & 0.2894 & 0.5344 & 0.5946 & 0.5031  \\
\textbf{SimAC} & 0.2398 & 0.3390 & 0.3021 & 0.4049 & 0.5131 & 0.4339 & 0.2499 & 0.3240 & 0.2849 & 0.4483 & 0.5262 & 0.4455  \\
\textbf{AdvDM} & 0.2827 & 0.3907 & 0.3125 & 0.4158 & 0.5007 & 0.4464 & 0.2222 & 0.3484 & 0.2680 & 0.5056 & 0.5435 & 0.4669  \\
\textbf{ACE} & 0.3462 & 0.4354 & 0.3505 & 0.5412 & 0.6673 & 0.5834 & 0.2418 & 0.3422 & 0.2880 & 0.4813 & 0.5480 & 0.4839  \\
\textbf{IDProtector} & 0.2543 & 0.3071 & 0.2922 & 0.4363 & 0.5421 & 0.4486 & 0.1881 & \textbf{0.3003} & \textbf{0.2256} & 0.5065 & 0.6161 & 0.5611  \\
\textbf{IDGuar (Ours)} & \textbf{0.1003} & \textbf{0.0515} & \textbf{0.0940} & \textbf{0.2686} & \textbf{0.2905 }& \textbf{0.2385} & \textbf{0.1794} & 0.3061 & 0.2431 & \textbf{0.3469} & \textbf{0.4198 }&\textbf{ 0.3489} \\
\specialrule{1.2pt}{0pt}{0pt}
 \multirow{2}{*}{\textbf{Method}}  & \multicolumn{3}{c}{\textbf{\textbf{\textbf{Photomaker}}}} & \multicolumn{3}{c}{\textbf{\textbf{\textbf{Dreambooth}}}} & \multicolumn{3}{c}{\textbf{\textbf{\textbf{InfiniteYou}}}} & \multicolumn{3}{c}{\textbf{\textbf{\textbf{Mean}}}}  \\
\cmidrule(lr){2-4} \cmidrule(lr){5-7} \cmidrule(lr){8-10} \cmidrule(lr){11-13} 
&\textbf{Arc $\downarrow$} & \textbf{Face $\downarrow$} & \textbf{VGG $\downarrow$}&\textbf{Arc $\downarrow$} & \textbf{Face $\downarrow$} & \textbf{VGG $\downarrow$} &\textbf{Arc $\downarrow$} & \textbf{Face $\downarrow$} & \textbf{VGG $\downarrow$}&\textbf{Arc $\downarrow$} & \textbf{Face $\downarrow$} & \textbf{VGG $\downarrow$} \\
\specialrule{1.0pt}{0pt}{0pt}
\textbf{No Protected}  & 0.6934 & 0.4219 & 0.2612 & 0.5421 & 0.6555& 0.4335& 0.6614 & 0.7485 & 0.5640 & 0.5428 & 0.5787 & 0.4445\\
\textbf{Anti-DB}    & 0.6504 & 0.4065 & 0.2560 &0.1031 &0.0666& 0.0761& 0.6035 & 0.6869 & 0.5312  & 0.4167 & 	0.4412 & 0.3573\\
\textbf{SimAC}   & 0.6251 & 0.3564 & 0.2321 &\textbf{0.0831}&0.0852&0.0665 & 0.5854 & 0.6846 & 0.4654 & 0.3766 & 0.4041 & 0.3186\\
\textbf{AdvDM}  & 0.6313 & 0.3718 & 0.2672&0.0868&0.0365&0.1100& 0.5207 & 0.6293 & 0.4654 & 0.3807 & 0.4030 & 0.3338\\
\textbf{ACE}   & 0.6422 & 0.3880 & 0.2384 & 0.1401& 0.2271&0.0944& 0.6358 & 0.7283 & 0.5560 & 0.4327 & 0.4766 & 0.3707 \\
\textbf{IDProtector} & 0.6151 & 0.3013 & 0.2022 & 0.3298 & 0.3727 & 0.2578 & 0.5617 & 0.6609 & 0.4943 & 0.4131 & 0.4429 & 0.3545  \\
\textbf{IDGuar (Ours)}  & \textbf{0.3320} & \textbf{0.2654} & \textbf{0.1996} & 0.0838 &\textbf{0.0320} & \textbf{0.0610} & \textbf{0.3331} & \textbf{0.3871} & \textbf{0.2428} & \textbf{0.2349} & \textbf{0.2503} & \textbf{0.2040}\\
\specialrule{1.2pt}{0pt}{0pt}
\end{tabular}

\label{tab:512 performance_comparison}
\end{table*}

\begin{table*}[htbp]
\centering
\caption{Comparison of identity protection performance against baseline methods on the CelebA-HQ dataset. Evaluation is conducted using FaceQNet and MagFace to measure the perceptual quality of generated images.}
\renewcommand{\arraystretch}{1.5}
\setlength{\tabcolsep}{1.06mm}
\begin{tabular}{lccccccccccccccccccc}
\specialrule{1.2pt}{0pt}{0pt}
    \multirow{2}{*}{\textbf{Method}} & \multicolumn{2}{c}{\textbf{Ip-Adapter}} & \multicolumn{2}{c}{\textbf{Ip-Adapter Plus XL}} & \multicolumn{2}{c}{\textbf{Blip-Diffusion}} & \multicolumn{2}{c}{\textbf{InfiniteID}}  \\
\cmidrule(lr){2-3} \cmidrule(lr){4-5} \cmidrule(lr){6-7} \cmidrule(lr){8-9} 
& \textbf{FaceQNet $\downarrow$} & \textbf{MagFace $\downarrow$} & \textbf{FaceQNet $\downarrow$} & \textbf{MagFace $\downarrow$} &\textbf{FaceQNet $\downarrow$} & \textbf{MagFace $\downarrow$} & \textbf{FaceQNet $\downarrow$} & \textbf{MagFace $\downarrow$}\\
\specialrule{1.0pt}{0pt}{0pt}
\textbf{No Protected}       & 0.3031 & 20.6785 & 0.4490 & 18.8023 & 0.3344 & 20.9884 & 0.4527 & 19.6607 \\
\textbf{Anti-DB}    & 0.2599 & \textbf{19.5696} & 0.4515 & \textbf{18.5281} & 0.3170 & 21.4418 & 0.4731 & 18.9686 \\
\textbf{SimAC}      & 0.2670 & 19.9800 & 0.4493 & 19.2084 & 0.3433 & \textbf{19.4777} & 0.4695 & 19.0426 \\
\textbf{AdvDM}      & 0.2842 & 19.9800 & 0.4505 & 19.2084 & \textbf{0.2941} & 20.0386 & 0.4645 & 19.2935 \\
\textbf{ACE}        & 0.2928 & 20.0025 & 0.4540 & 19.3925 & 0.3363 & 20.7060 & 0.4539 & 19.1724 \\
\textbf{IDProtector}        & 0.2945 & 20.0540 & 0.4544 & 19.0985 & 0.3039 & 20.4320 & \textbf{0.4204} & 19.7844 \\
\textbf{IDGuar (Ours)} & \textbf{0.2556} & 19.9783 & \textbf{0.4490} & 18.7536 & 0.3098 & 20.6180 & 0.4574 & \textbf{18.7834} \\
\specialrule{1.2pt}{0pt}{0pt}
\multirow{2}{*}{\textbf{Method}} & \multicolumn{2}{c}{\textbf{Photomaker}} & \multicolumn{2}{c}{\textbf{Dreambooth}} & \multicolumn{2}{c}{\textbf{InfiniteYou}} & \multicolumn{2}{c}{\textbf{Mean}}  \\
\cmidrule(lr){2-3} \cmidrule(lr){4-5} \cmidrule(lr){6-7} \cmidrule(lr){8-9} 
& \textbf{FaceQNet $\downarrow$} & \textbf{MagFace $\downarrow$} & \textbf{FaceQNet $\downarrow$} & \textbf{MagFace $\downarrow$} &\textbf{FaceQNet $\downarrow$} & \textbf{MagFace $\downarrow$} & \textbf{FaceQNet $\downarrow$} & \textbf{MagFace $\downarrow$}  \\
\specialrule{1.0pt}{0pt}{0pt}
\textbf{No Protected}       & 0.2674 & 20.5900 & 0.3222 & 21.1998 & 0.4463 & 19.4956 & 0.3679 & 20.2022 \\
\textbf{Anti-DB}    & 0.2693 & \textbf{20.4309} & 0.2802 & 21.8453 & 0.4451 & 19.6502 & 0.3566 & 20.0621 \\
\textbf{SimAC}      & 0.2693 & 20.5074 & 0.2650 & 21.6122 & 0.4489 & 19.7678 & 0.3589 & 19.9423 \\
\textbf{AdvDM}      & 0.2672 & 20.5074 & 0.3072 & 21.3755 & 0.4418 & 19.7678 & 0.3585 & 20.0245 \\
\textbf{ACE}        & 0.2677 & 20.7290 & \textbf{0.2501} & 21.5659 & 0.4463 & 19.4555 & 0.3573 & 20.1463 \\
\textbf{IDProtector}        & 0.2907 & 28.8511 & 0.3740 & 20.0474 & 0.4565 & 21.7575 & 0.3706 & 21.4321  \\
\textbf{IDGuar (Ours)} & \textbf{0.2653} & 20.5549 & 0.3073 & \textbf{21.2664} & \textbf{0.4352} & \textbf{19.3972} & \textbf{0.3542} & \textbf{19.9074} \\
\specialrule{1.2pt}{0pt}{0pt}
\end{tabular}
\label{tab:512 Performance.}
\end{table*}

\begin{table*}[t]
\centering

  \caption{ ArcFace-only identity similarity on LFW. Method names are abbreviated for brevity.}
\label{tab:R3_lfw_arc_only}
\renewcommand{\arraystretch}{1.5}
\setlength{\tabcolsep}{1.3mm}
\begin{tabular}{lcccccccc}
\specialrule{1.2pt}{0pt}{0pt}
\textbf{Method}
& \textbf{IP-Adapter$\downarrow$} 
& \textbf{IP-Ada. + XL$\downarrow$}
& \textbf{BLIP-D.$\downarrow$}
& \textbf{InfiniteID$\downarrow$}
& \textbf{PhotoM.$\downarrow$}
& \textbf{DreamB.$\downarrow$}
& \textbf{InfiniteYou$\downarrow$}
& \textbf{Mean$\downarrow$} \\
\specialrule{1.2pt}{0pt}{0pt}
\textbf{No Protected}   & 0.3580 & 0.5210 & 0.3010 & 0.6150 & 0.3420 & 0.4550 & 0.6670 & 0.4656 \\
\textbf{Anti-DB}        & 0.2550 & 0.4890 & 0.2490 & 0.5720 & 0.2300 & 0.3380 & 0.5980 & 0.3901 \\
\textbf{SimAC}          & 0.2070 & 0.3620 & 0.2760 & 0.5490 & 0.2200 & 0.2870 & 0.5570 & 0.3511 \\
\textbf{AdvDM}          & 0.2320 & 0.3480 & 0.2410 & 0.5030 & 0.2050 & 0.2540 & 0.5410 & 0.3320 \\
\textbf{ACE}            & 0.2480 & 0.4620 & 0.2610 & 0.5330 & 0.2270 & 0.3210 & 0.6400 & 0.3846 \\
\textbf{IDProtector}    & 0.2130 & 0.3310 & 0.2380 & 0.4920 & 0.2180 & 0.3160 & 0.5090 & 0.3310 \\
\textbf{IDGuar (Ours)}& \textbf{0.0790} & \textbf{0.1360} & \textbf{0.1980} & \textbf{0.4050} &
                          \textbf{0.0910} & \textbf{0.2520} & \textbf{0.3180} & \textbf{0.2113} \\
\specialrule{1.2pt}{0pt}{0pt}
\end{tabular}
\end{table*}

\begin{figure*}[htbp]
    \centering
    \includegraphics[width=1\textwidth]{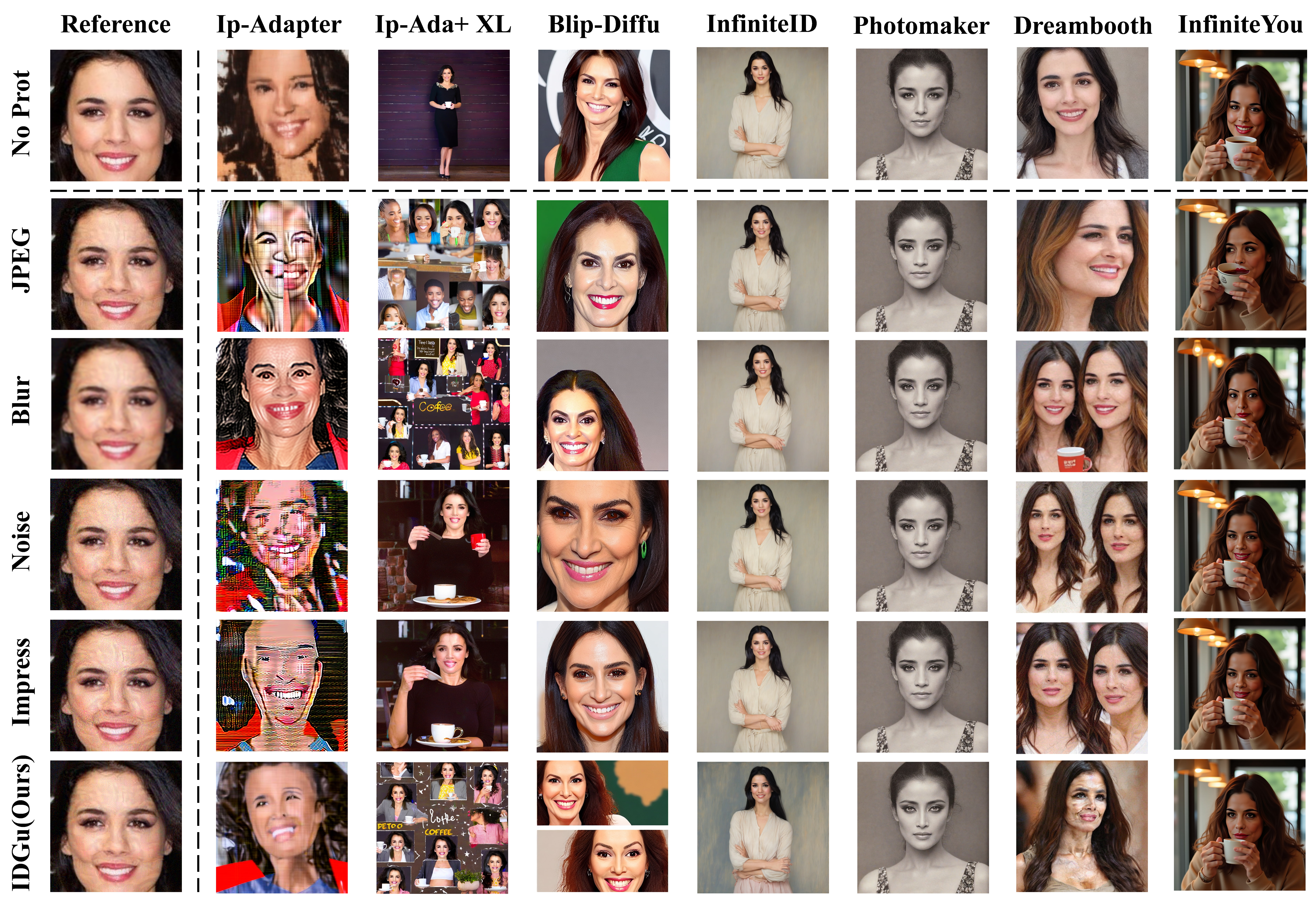}
    \caption{Robustness Evaluation Against Image Post-Processing and Defense Attacks.}
    \label{fig:vis2}
\end{figure*}

\begin{figure*}[htbp]
    \centering
    \includegraphics[width=1\textwidth]{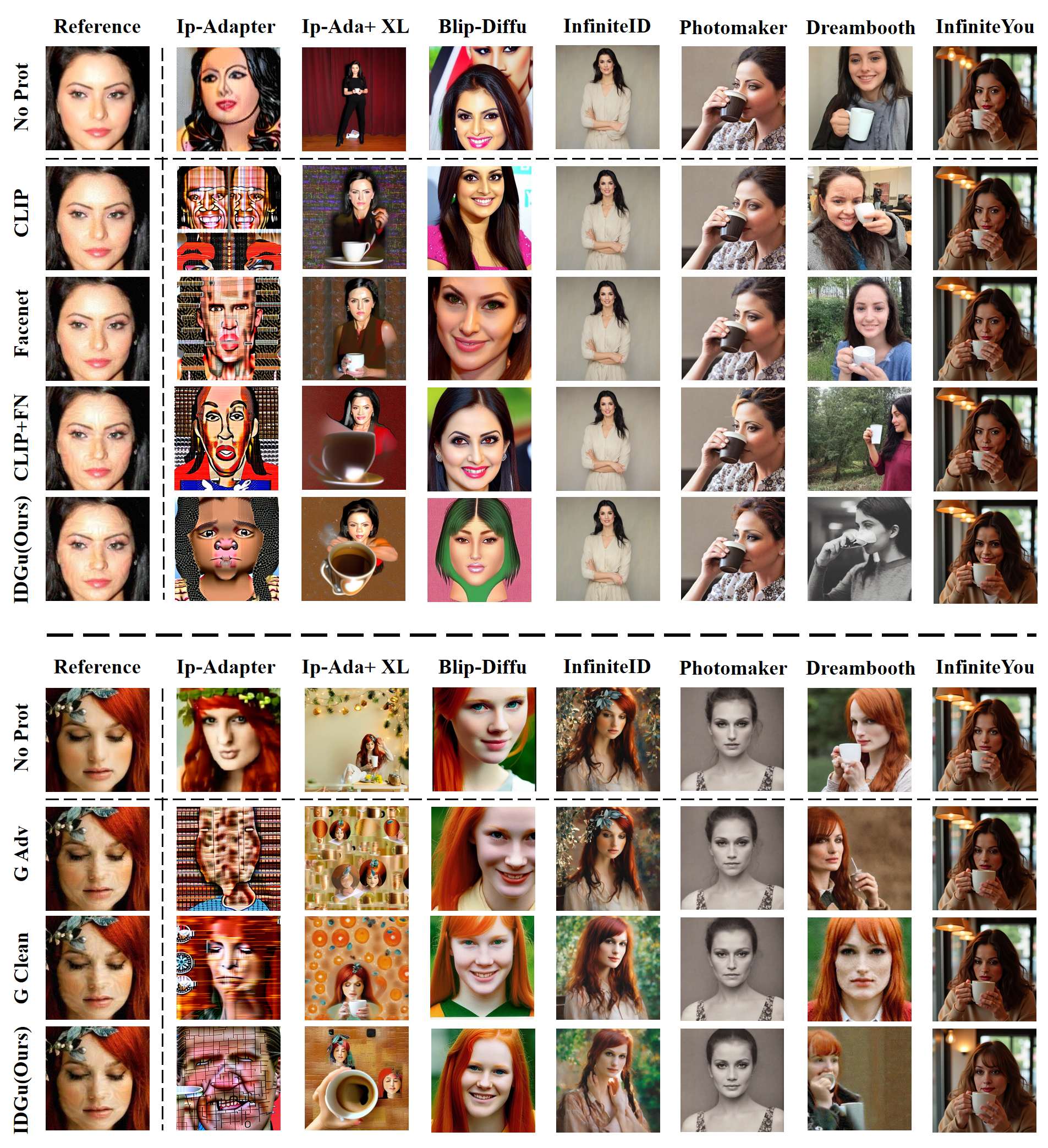}
    \caption{Visualization of the ablation study on identity loss combinations and guidance directions.}
    \label{fig:vis4}
\end{figure*}

\begin{figure*}[htbp]
    \centering
    \includegraphics[width=1\textwidth]{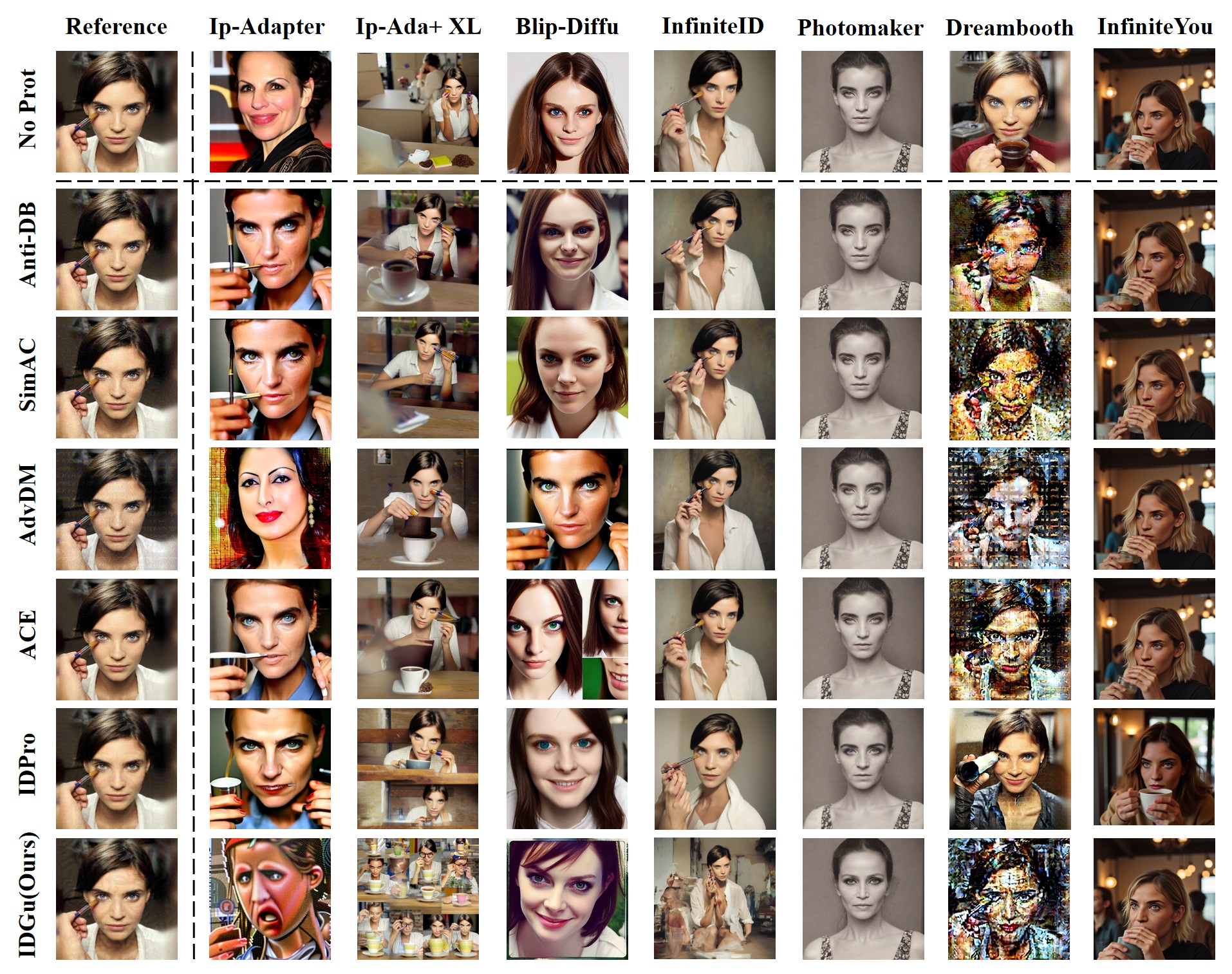}
    \caption{Visualization of Identity Suppression Effectiveness on CelebA-HQ Dataset for IDGuardian and Comparison Methods.}
    \label{fig:vis3}
\end{figure*}



\end{document}